\documentclass[10pt,journal,compsoc]{IEEEtran}
\usepackage{times}
\usepackage{epsfig}
\usepackage{graphicx}
\usepackage{amsmath}
\usepackage{amssymb}

\usepackage{caption}
\usepackage{subcaption}
\usepackage{epsfig}
\usepackage{amsmath}
\usepackage{amssymb}
\usepackage{booktabs}
\usepackage{rotating}
\usepackage{algorithm} 
\usepackage{algpseudocode}
\usepackage{multirow}
\usepackage{color}

\usepackage{enumitem}
\usepackage{enumerate}

\newcommand{\etal}{\textit{et al}.}
\newcommand{\ie}{\textit{i}.\textit{e}.}
\newcommand{\eg}{\textit{e}.\textit{g}.}

\newif\ifshowcomments
\showcommentstrue

\ifshowcomments

\newcommand{\TODO}[1]{{\color{red}{[TODO: #1]}}}
\newcommand{\revised}[1]{{\color[rgb]{0.0,0.0,0.0}{#1}}}
\newcommand{\wty}[1]{{\color[rgb]{0.0,0.0,0.0}{#1}}}
\newcommand{\Phil}[1]{{\color[rgb]{0.0,0.0,0.0}{#1}}}
\newcommand{\phil}[1]{{\color[rgb]{0.0,0.0,0.0}{#1}}}
\newcommand{\xwhu}[1]{{\color[rgb]{0.0,0.0,0.0}{#1}}}
\newcommand{\hxw}[1]{{\color[rgb]{0.0,0.0,0.0}{#1}}}
\else
\newcommand{\TODO}[1]{}
\newcommand{\revised}[1]{}
\newcommand{\wty}[1]{}
\newcommand{\Phil}[1]{}
\newcommand{\xwhu}[1]{}
\newcommand{\hxw}[1]{}
\newcommand{\phil}[1]{}\
\fi

\ifCLASSOPTIONcompsoc
  \usepackage[nocompress]{cite}
\else
  \usepackage{cite}
\fi
\ifCLASSINFOpdf
\else
\fi
\usepackage[pagebackref=false,breaklinks=true,letterpaper=true,colorlinks,citecolor=black,linkcolor=black,bookmarks=false]{hyperref}

\begin{document}
%
\title{Instance Shadow Detection \\ with A Single-Stage Detector }
%
%
%

\author{Tianyu Wang, Xiaowei~Hu*, Pheng-Ann Heng, and Chi-Wing~Fu
    \IEEEcompsocitemizethanks{
	\IEEEcompsocthanksitem T. Wang and C.-W. Fu are with the Department of Computer Science and Engineering, The Chinese University of Hong Kong and the Shun Hing Institute of Advanced Engineering, The Chinese University of Hong Kong.
	\IEEEcompsocthanksitem X. Hu is with the Shanghai AI Laboratory.
	 	\IEEEcompsocthanksitem P.-A. Heng is with the Department of Computer Science and Engineering, The Chinese University of Hong Kong.
	\IEEEcompsocthanksitem The preliminary versions of this work were accepted for presentation in CVPR 2020~\cite{wang2020instance} and CVPR 2021~\cite{wang2021single}. 
    \IEEEcompsocthanksitem Corresponding author: Xiaowei Hu (huxiaowei@pjlab.org.cn).
}
}

\markboth{IEEE Transactions on Pattern Analysis and Machine Intelligence}%
{Shell \MakeLowercase{\textit{et al.}}: Bare Demo of IEEEtran.cls for Computer Society Journals}
\IEEEtitleabstractindextext{%

\begin{abstract}
This paper formulates a new problem, instance shadow detection, which aims to detect shadow instance and the associated object instance that cast each shadow in the input image. 
To approach this task, we first compile a new dataset with the masks for shadow instances, object instances, and shadow-object associations.
We then design an evaluation metric for quantitative evaluation of the performance of instance shadow detection.
Further, we design a single-stage detector to perform instance shadow detection in an end-to-end manner, where the bidirectional relation learning module and the deformable maskIoU head are proposed in the detector to directly learn the relation between shadow instances and object instances and to improve the accuracy of the predicted masks.
Finally, we quantitatively and qualitatively evaluate our method on the benchmark dataset of instance shadow detection and show the applicability of our method on light direction estimation and photo editing.
%
%
\end{abstract}

\begin{IEEEkeywords}
Instance shadow detection, instance segmentation, shadow detection, deep neural network.
\end{IEEEkeywords}}

\maketitle

\IEEEdisplaynontitleabstractindextext

%
\IEEEpeerreviewmaketitle

\section{Introduction}
\label{sec:introduction}

\noindent
``{\em When you light a candle, you also cast a shadow,\/}''---Ursula K. Le Guin written in A Wizard of Earthsea.

Shadows are formed when the light is blocked by the objects.
When we see a shadow, we also know that there must be some objects that create or cast the shadow.
However, recent shadow detection methods~\cite{hou2021large,hu2021revisiting,Hu_2018_CVPR,DBLP:journals/pami/HuFZQH20,khan2014automatic,le2018a+d,khan2016automatic,vicente2016large,zheng2019distraction,zhu2018bidirectional,chen2020multi} simply generate a binary mask to indicate the shadow regions and fail to find the associated object that casts each individual shadow.
To {\em find shadows together with their associated objects\/}, we propose a new task, named {\em instance shadow detection\/}, in which we detect not only individual shadow instances in the input image but also the associated object that casts each shadow.

Instance shadow detection has the potential to benefit various applications.
For privacy protection, for example, when we remove vehicles and persons from photos, we can remove the associated shadows with the objects together.
%
%
Also for photo editing, when we translate or scale objects in photos, we can naturally manipulate the objects with their associated shadows simultaneously.
Further, shadow-object associations give hints to estimate the light direction, facilitating the development of applications such as shadow generation for virtual objects in AR environments and scene relighting.
Last, the detected shadow and object instances help to estimate building heights from satellite metadata~\cite{hao2021building}. 

\begin{figure}[tp]
	\centering
	\includegraphics[width=0.98\linewidth]{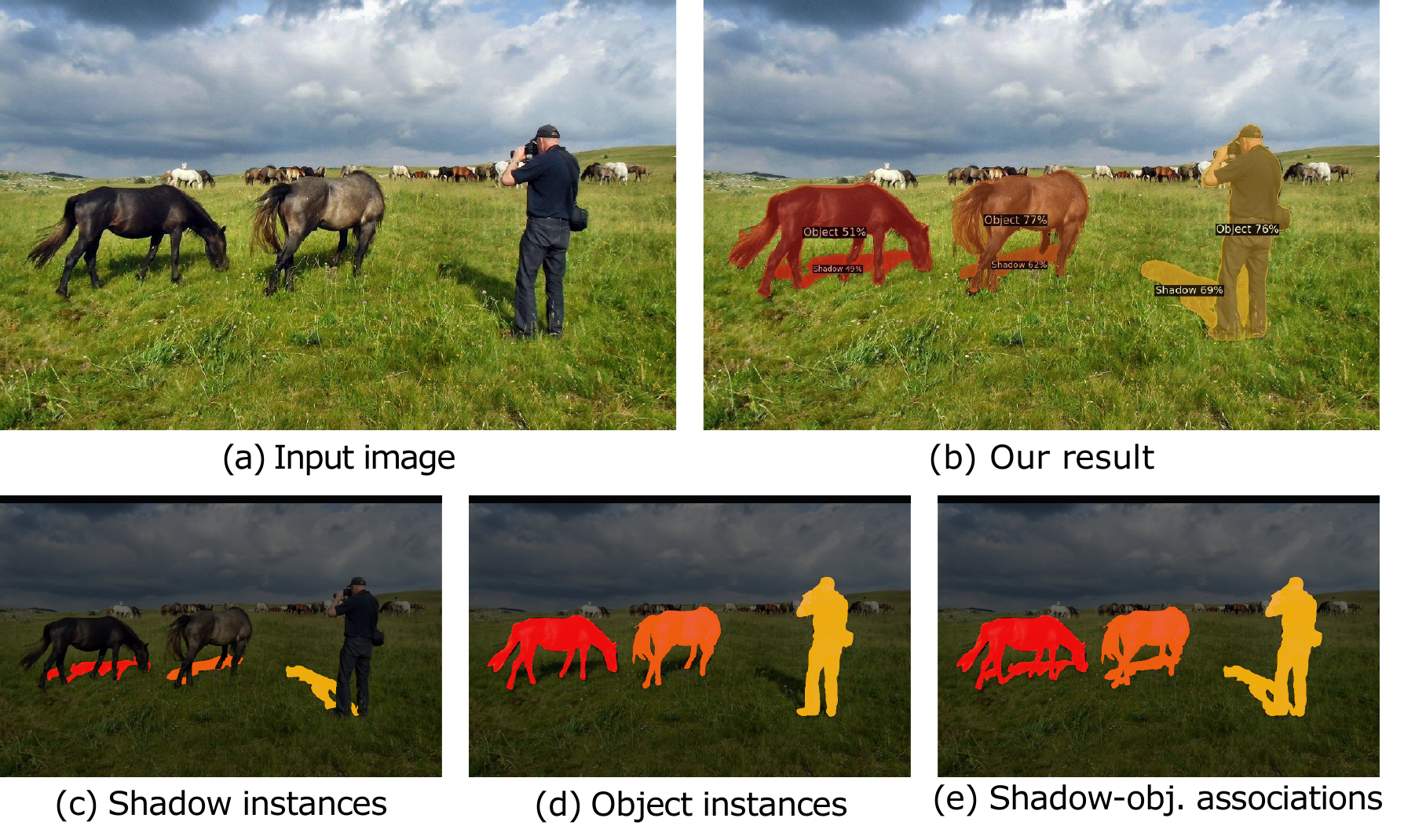}
	\caption{
		Given (a) an input photo, the goal of instance
		shadow detection is to detect (c) individual shadow instances, 
		(d) individual object instances , and (e) shadow-object associations. (b) shows the overall result produced by our method from (a).} 
	\label{img:1}
	\vspace{-6mm}
\end{figure}

%
\wty{To approach the new task of instance shadow detection, we first prepared the SOBA (Shadow OBject Association) dataset.
The dataset has three parts: SOBA training, SOBA testing, and SOBA challenge. 
Both the SOBA-testing and SOBA-challenge sets are for testing but the SOBA-challenge set contains complex scenarios for evaluating the capability of methods in handling challenging cases. 
The whole dataset contains 4,293 pairs of annotated shadow-object associations over 1,100 images.}
%
Each image has (i) a shadow instance mask, which labels each shadow instance with a unique color; (ii) a shadow-object association mask, which labels each shadow-object pair with a corresponding unique color; and (iii) an object instance mask, which is (ii) minus (i); see Figure~\ref{img:1} for an example. 
%
Also, we formulate a new evaluation metric for the task to quantitatively evaluate the performance of the instance shadow detection results.

We approach the instance shadow detection task by exploiting the remarkable computational capability of deep neural networks.
%
%
\wty{Our earlier work LISA~\cite{wang2020instance}} first generates region proposals that likely contain shadow/object instances and shadow-object associations.
For each proposal, we then crop regions-of-interest (RoIs) from the feature maps and predict masks and boxes of the shadow instances, object instances, and shadow-object associations from each RoI. 
%
Lastly, we pair the shadow and object instances with the shadow-object associations.
However, this two-stage framework and post-processing strategy have several limitations.
%
%
First, this approach considers shadow-object association as a single category. Yet, the appearance of shadow and object instances have large variations, so shadow-object associations could easily be missed.
Second, it generates region proposals for shadow/object instances and shadow-object associations using two separate branches and leverages post-processing to produce the final shadow-object associations.
Errors could accumulate over the network and post-processing.
%
Third, the employed RoIs represent feature regions using rectangular shapes. However, the shapes of the shadow instances and shadow-object associations are usually irregular and the cropped RoIs of rectangular shapes could include many irrelevant image contents such as other object and shadow instances.

To address the above issues, we design a single-stage deep framework~\cite{wang2021single} 
to directly learn to find the relation between shadow and object instances in an end-to-end manner.
%
%
This framework includes only fully convolutional operations to generate the masks for the shadow instances, object instances,
and shadow-object associations, thus enabling us to handle shadow/object instances and shadow-object associations of any shape. 
Importantly, we design the bidirectional relation learning module to find the shadow-object association pairs to learn an offset vector from the center
of each shadow instance to the center of its associated object instance, and the other way around, aiming to explore the inter-relationship between shadows and objects effectively. 
We construct a class vector to represent the learning directions during this process: shadow to object or object to shadow.
%

Further, we design a deformable MaskIoU head in the network to improve the mask accuracy.
This module takes the output of the mask head and mask feature as inputs and predicts the IoU scores of the predicted masks. 
%
\wty{Unlike the Mask Scoring R-CNN~\cite{huang2019msrcnn}, which feeds feature cropped from the RoI as the input of the MaskIOU head, we take the raw feature as input to our method.
Hence, we further introduce the deformable convolution~\cite{zhu2019defor} to process the whole feature and focus on discriminate regions in the shadow/object instance masks.}
%
Also, we formulate a segmentation loss, an offset loss, and a boundary loss to jointly optimize the entire network. 
Lastly, we design also a shadow-aware copy-and-paste strategy to augment input images during the training.
These new techniques help the network learn to better pair the shadow and object instances for improving the overall performance.


Below, we summarize the major contributions of this work.
\begin{itemize}[]
\vspace*{-0.25mm}
\item
First, we formulate a new task, instance shadow detection, which aims to find individual shadow instances, individual object instances, and the shadow-object associations.
\item
Second, we prepare a new dataset and evaluation metric to support instance shadow detection. The dataset contains 1,100 images and 4,293 pairs of shadow-object associations, and provides three instance masks for each image.
\item
Third, we design a single-stage instance shadow detection network with two novel techniques, the bidirectional relation learning module, the deformable maskIoU head, and some training strategies to directly learn the relation between shadow and object instances.
%
\item
Fourth, we perform various experiments to quantitatively and visually demonstrate the effectiveness of our method. Results show that our method outperforms our previous two-stage detector~\cite{wang2020instance} by over $50.2\%$ and $70.2\%$ on the SOBA-testing set and the SOBA-challenge set, respectively.
\item
Last, we demonstrate the applicability of the instance shadow detection on various tasks, including light direction estimation and photo editing.
\end{itemize}

\begin{figure*}[tp]     
	\centering
	\begin{minipage}[t]{\textwidth}
		\centering
		\includegraphics[width=0.97\linewidth]{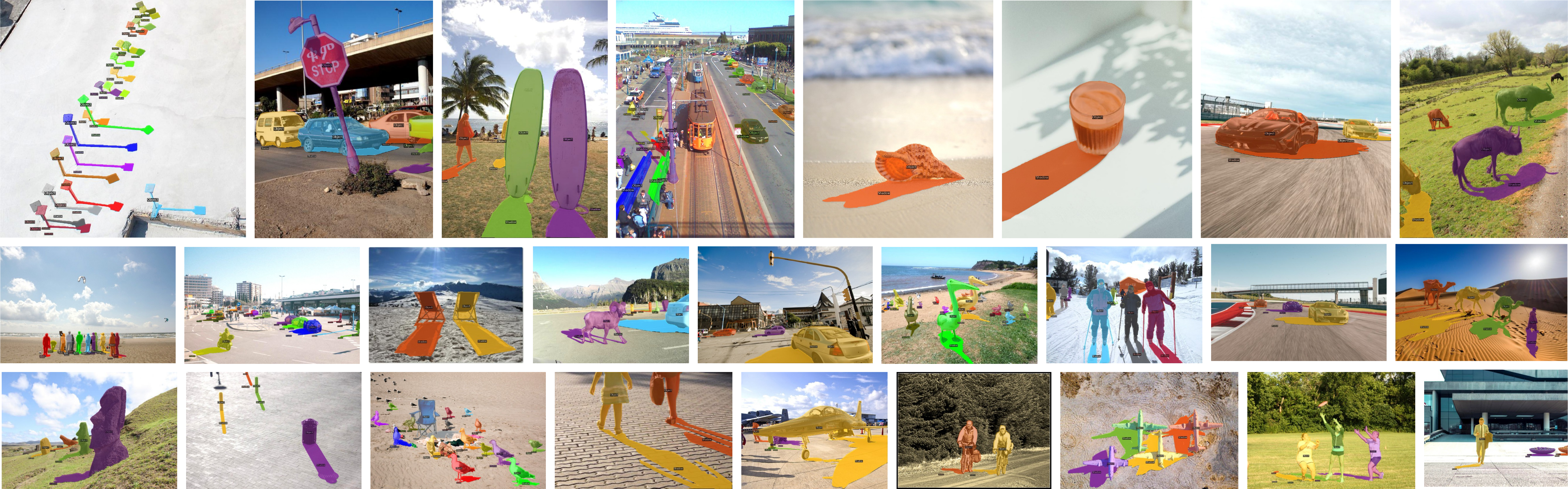}
		\centerline{\footnotesize (a) Example images in the SOBA-training set}
		\vspace*{0.5mm}
	\end{minipage}
	\begin{minipage}[t]{0.495\textwidth}
		\centering
		\includegraphics[width=0.96\linewidth]{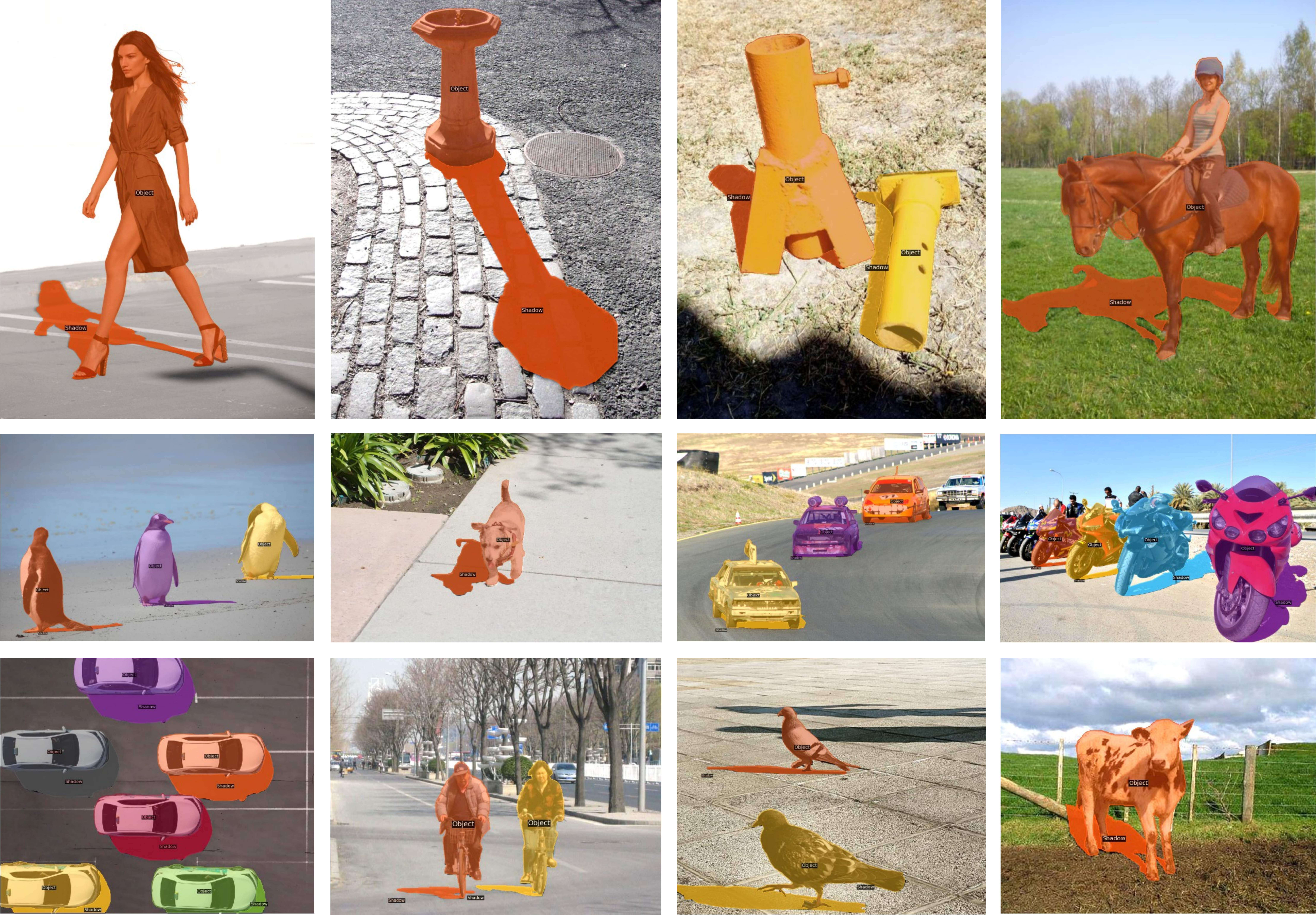}
		\centerline{\footnotesize (b) Example images in the SOBA-testing set}
	\end{minipage}
	\begin{minipage}[t]{0.495\textwidth}
		\centering
		\includegraphics[width=0.96\linewidth]{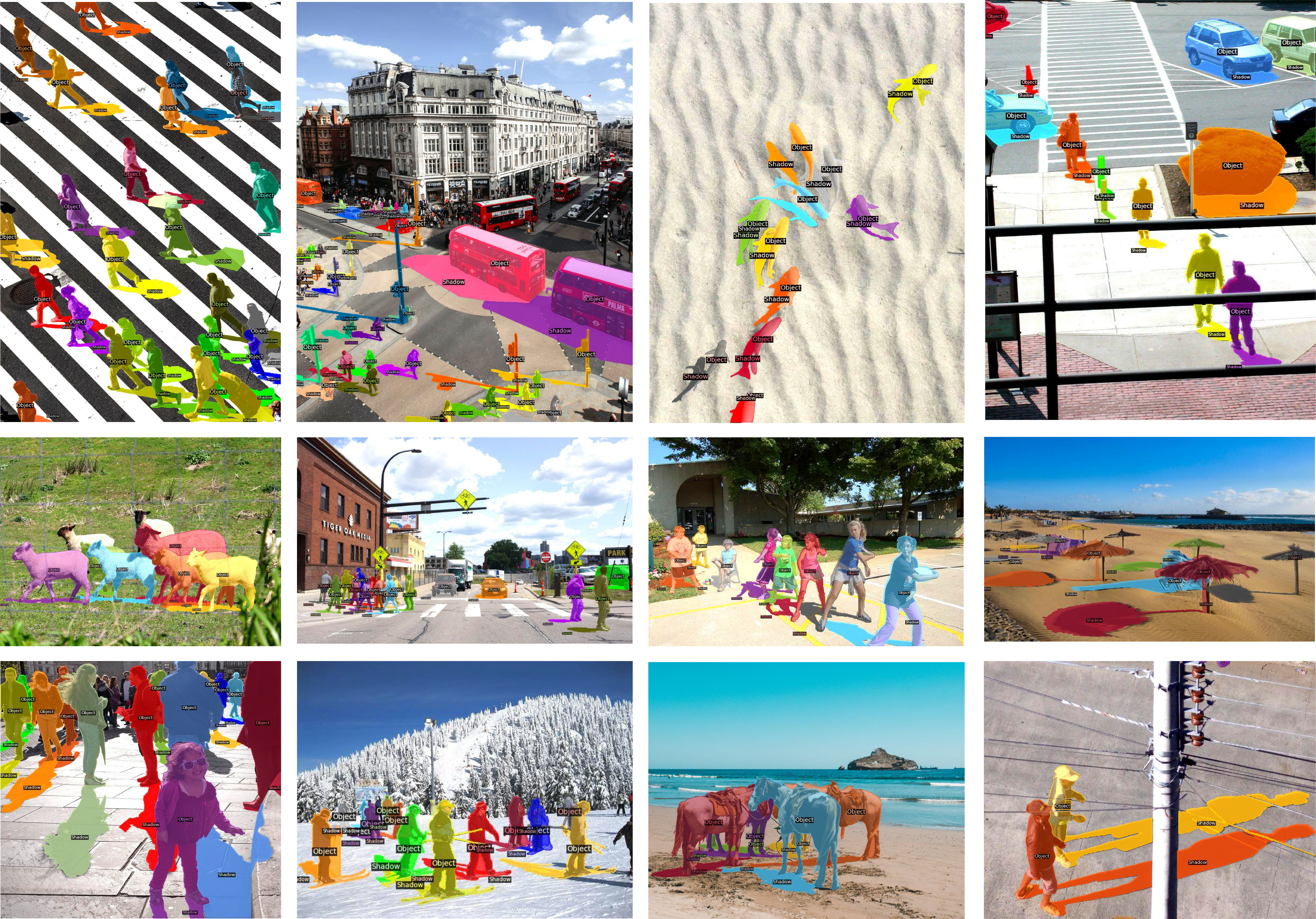}
		\centerline{\footnotesize (c) Example images in the SOBA-challenge set}
	\end{minipage}
	\vspace*{-1mm}
	\caption{Example images with the mask labels in our SOBA data set. Please zoom in for a better visualization.}
	\label{img:dataset_SOBA}
	\vspace*{-3mm}
\end{figure*}

\section{Related Work}
\label{sec:related}

\noindent
{\bf Shadow detection.} \
Generic shadow detection aims to generate a binary mask to mark shadow regions in the input image.
Early methods build physical models to leverage color and illumination to detect shadows.
Among them, Salvador~\etal~\cite{salvador2004cast} explore shadows' spectral and geometrical properties to segment the cast shadows. 
Panagopoulos~\etal~\cite{panagopoulos2011illumination} build a higher-order Markov Random Field illumination model with coarse 3D geometry information.
Tian~\etal~\cite{tian2016new} adopt the difference of spectral power distributions in daylight and skylight for shadow detection.

Later, machine-learning approaches are developed to recognize shadows by first describing image regions using hand-crafted features and then classifying the regions into shadows and non-shadows.
Features like texture~\cite{zhu2010learning,vicente2015leave,guo2011single,vicente2018leave},  T-junction~\cite{lalonde2010detecting}, color~\cite{lalonde2010detecting,vicente2015leave,guo2011single,vicente2018leave}, and  edge~\cite{lalonde2010detecting,zhu2010learning,huang2011characterizes} are commonly used to describe shadows followed by classifiers like SVM~\cite{guo2011single,huang2011characterizes,vicente2015leave,vicente2018leave} and decision tree~\cite{lalonde2010detecting,zhu2010learning}.
These designed physical models and hand-crafted features have limited ability to describe shadows, so approaches based on these models and features may mis-detect shadows in general cases.

Deep neural networks automatically learn features from shadow images and show remarkable performance on shadow detection, especially with extensive training data. 
Khan~\etal~\cite{khan2014automatic} present the first work that uses a convolutional neural network (CNN) to learn features for shadow detection.
Shen~\etal~\cite{shen2015shadow} and Hou \& Vicente~\etal~\cite{hou2021large,vicente2016large} devise a structured learning framework and a stacked-CNN, respectively, to detect shadows. 
Nguyen~\etal~\cite{nguyen2017shadow} design an adjustable parameter in a conditional GAN to balance the weights of shadow and non-shadow regions.

Later, Hu~\etal~\cite{DBLP:journals/pami/HuFZQH20,Hu_2018_CVPR} learn the direction-aware spatial context to detect shadows by designing an attention mechanism in a spatial recurrent network.
%
Wang~\etal~\cite{wang2018stacked} iteratively detect and remove shadows with two conditional generative adversarial networks.
Le~\etal~\cite{le2018a+d} adopt adversarial training samples generated from a shadow attenuation network to train a shadow detection network. 
Zhu~\etal~\cite{zhu2018bidirectional} design a bidirectional feature pyramid network with recurrent attention residual modules to detect shadows.
Zheng~\etal~\cite{zheng2019distraction} revisit false negatives and false positives from the predicted results and derive a distraction-aware shadow detection network.
Ding~\etal~\cite{ding2019argan} detect and remove shadows in a recurrent manner via an attentive recurrent generative adversarial network.
More recently, Chen~\etal~\cite{chen2020multi} present a semi-supervised shadow detection algorithm by exploring unlabeled data through a multi-task mean teacher framework.
Hu~\etal~\cite{hu2021revisiting} build a new dataset to support shadow detection in a complex world and designed a fast shadow detection network.
Chen~\etal~\cite{chen2021triple} design a triple-cooperative video shadow detection network to detect shadows in videos.
Unlike general shadow detection, which adopts a single mask for all shadows in an image, this work detects not just individual shadows but also the associated objects altogether.

Apart from generic shadow detection, various works explored deep learning to remove shadows in natural images~\cite{khan2016automatic,qu2017deshadownet,hu2019mask,ding2019argan,Le_2019_ICCV,DBLP:conf/aaai/CunPS20,DBLP:conf/aaai/ZhangLZX20,le2020shadow,DBLP:journals/tip/LiuYMPW21,fu2021auto,liu2021shadow} and in documents~\cite{lin2020bedsr}, to generate shadows in augmented reality~\cite{liu2020arshadowgan} and in real scenes~\cite{hong2021shadow}, and to manipulate portrait shadows~\cite{DBLP:journals/tog/ZhangBTPZNJ20}.
Our instance shadow detection task offers a new perspective to edit or remove individual shadows with the associated objects.


\vspace*{3mm}  
\noindent
{\bf Instance segmentation.} \
Besides, this work relates to instance segmentation, which aims to label pixels of individual foreground objects in the input image.
One category of methods predicts region proposals in the input image and then generates an instance mask for each proposal,~\eg, 
MNC~\cite{dai2016instance}, DeepMask~\cite{pinheiro2015learning}, InstanceFCN~\cite{dai2016instance}, SharpMask~\cite{pinheiro2016learning}, FCIS~\cite{li2017fully}, BAIS~\cite{hayder2017boundary}, MaskLab~\cite{chen2018masklab}, Mask R-CNN~\cite{he2017mask}, PANet~\cite{liu2018path}, MegDet~\cite{peng2018megdet}, and HTC~\cite{chen2019hybrid}.
Among them, Mask R-CNN simultaneously predicts the category label, bounding box, and segmentation mask for each region proposal and achieves great success.
%
The other category directly predicts the instance masks and associated categories in the whole image,~\eg, TensorMask~\cite{chen2019tensormask}, SSAP~\cite{gao2019ssap}, SOLO~\cite{wang2020solo}, EmbedMask~\cite{ying2019embedmask}, SOLOv2~\cite{wang2020solov2}, CenterMask~\cite{lee2020centermask}, and CondInst~\cite{tian2020conditional}.
\wty{Our method is based on the architecture of CondInst~\cite{tian2020conditional}; below, we further elaborate on how CondInst works.}

\vspace*{3mm}
\noindent
{\bf \wty{Details on CondInst~\cite{tian2020conditional}.}}
\wty{
CondInst performs instance segmentation by generating the location and mask of each object. 
First, it adopts a fully convolutional network to predict the object centers based on the features extracted by the backbone network.
Second, it takes the object locations and the extracted mask features as inputs, and leverages the dynamic convolution to generate filters for each object to predict its mask. 
By doing so, CondInst can eliminate the RoI operations and reduce the parameters and computational complexity when predicting the masks, leading to a more efficient and simple instance segmentation framework.}
%
Based on CondInst, we further formulate our bidirectional relation learning module to learn the relation between shadow and object instances 
and design a deformable maskIoU head to penalize the predicted instance masks with low quality. 
%
%

\vspace*{3mm}
\noindent   
{\bf Difference from the conference papers.} \
This work extends our earlier works~\cite{wang2020instance,wang2021single} in three aspects.
First, we enrich our dataset prepared for instance shadow detection by providing more challenging cases with labels, aiming to evaluate the detection performance in complex scenarios.
%
Second, we improve the single-stage instance shadow detection (SSIS) method (in our conference version~\cite{wang2021single}) by designing new techniques:
(i) a deformable MaskIoU head,
(ii) a shadow-aware copy-and-paste data augmentation strategy, and 
(iii) a boundary loss, 
to better segment the shadow/object instances and shadow-object associations.
Our SSISv2 outperforms the two-stage detector LISA~\cite{wang2020instance} and the original SSIS~\cite{wang2021single} by 50.6\% and 17.2\%, respectively, in accuracy.
%
Third, we perform more experiments to evaluate the design of our network and add more applications to show how our SSISv2 outperforms the existing methods on instance shadow detection.


\section{Dataset and Evaluation Metric}

\begin{figure}[tp]     
\centering
\includegraphics[width=0.495\linewidth]{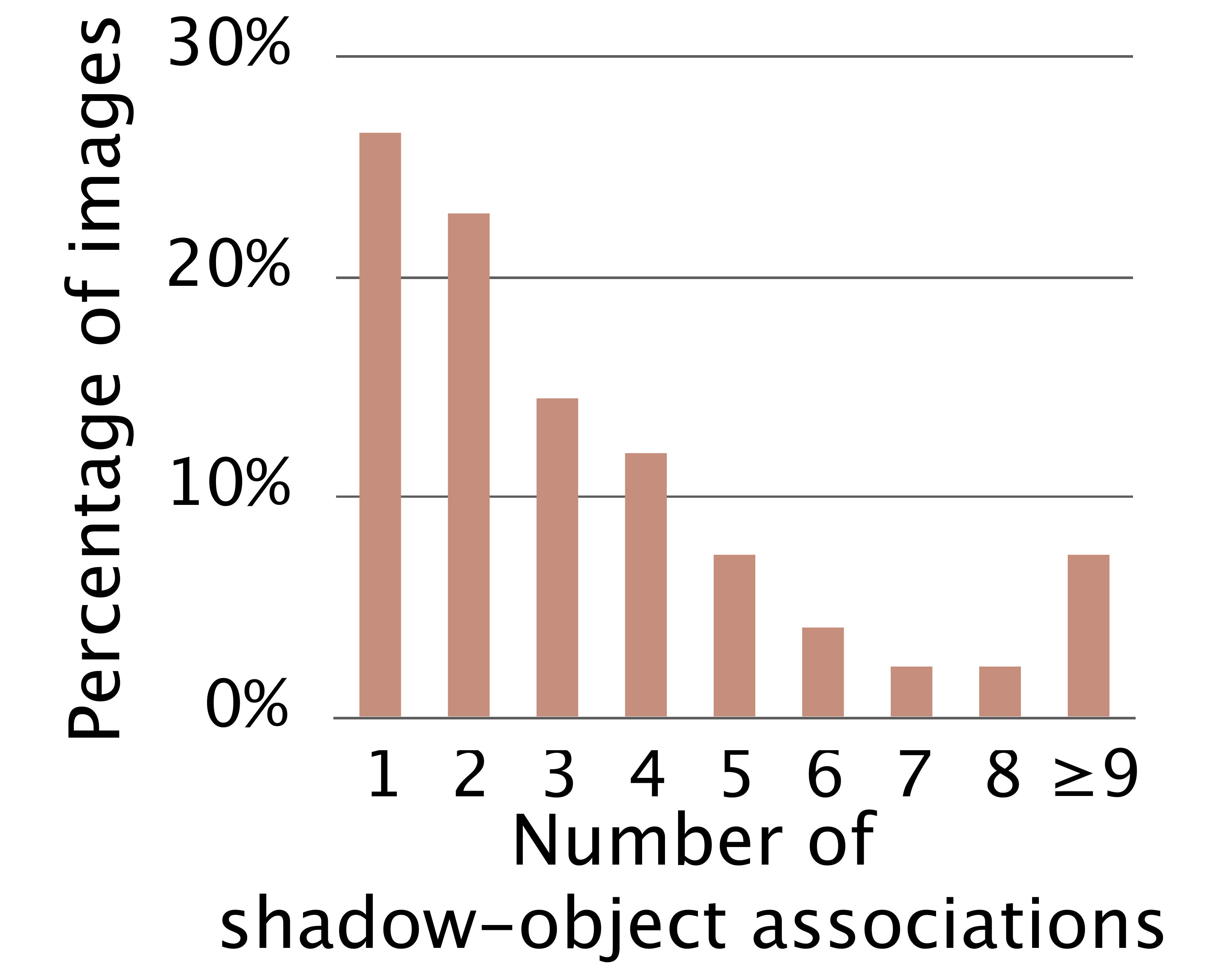}
\includegraphics[width=0.495\linewidth]{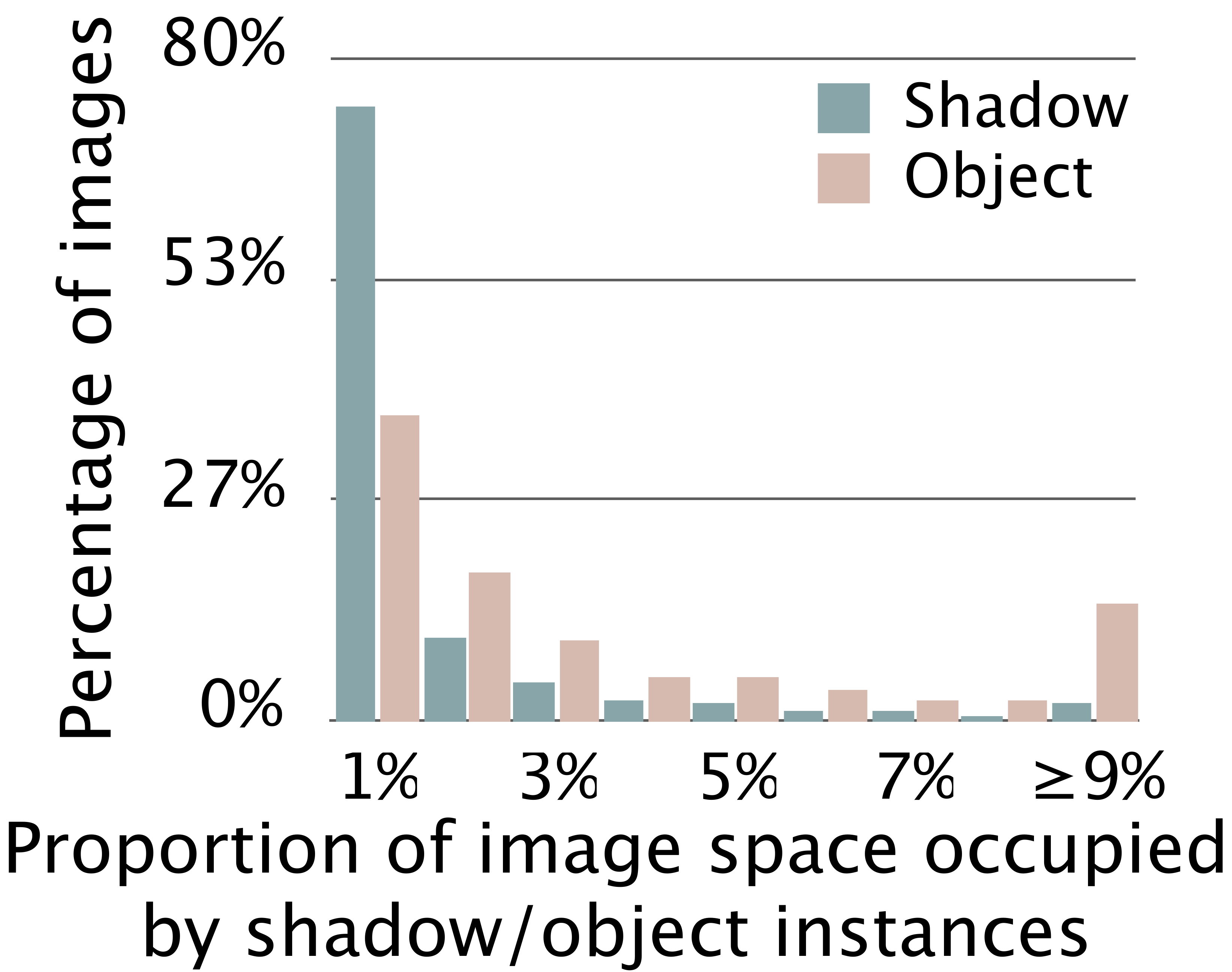}
\begin{minipage}[t]{\linewidth}
	\centering
	\centerline{\footnotesize (a) Statistical properties of the SOBA-training and SOBA-testing sets.}
	\vspace{4mm}
\end{minipage}
\includegraphics[width=0.495\linewidth]{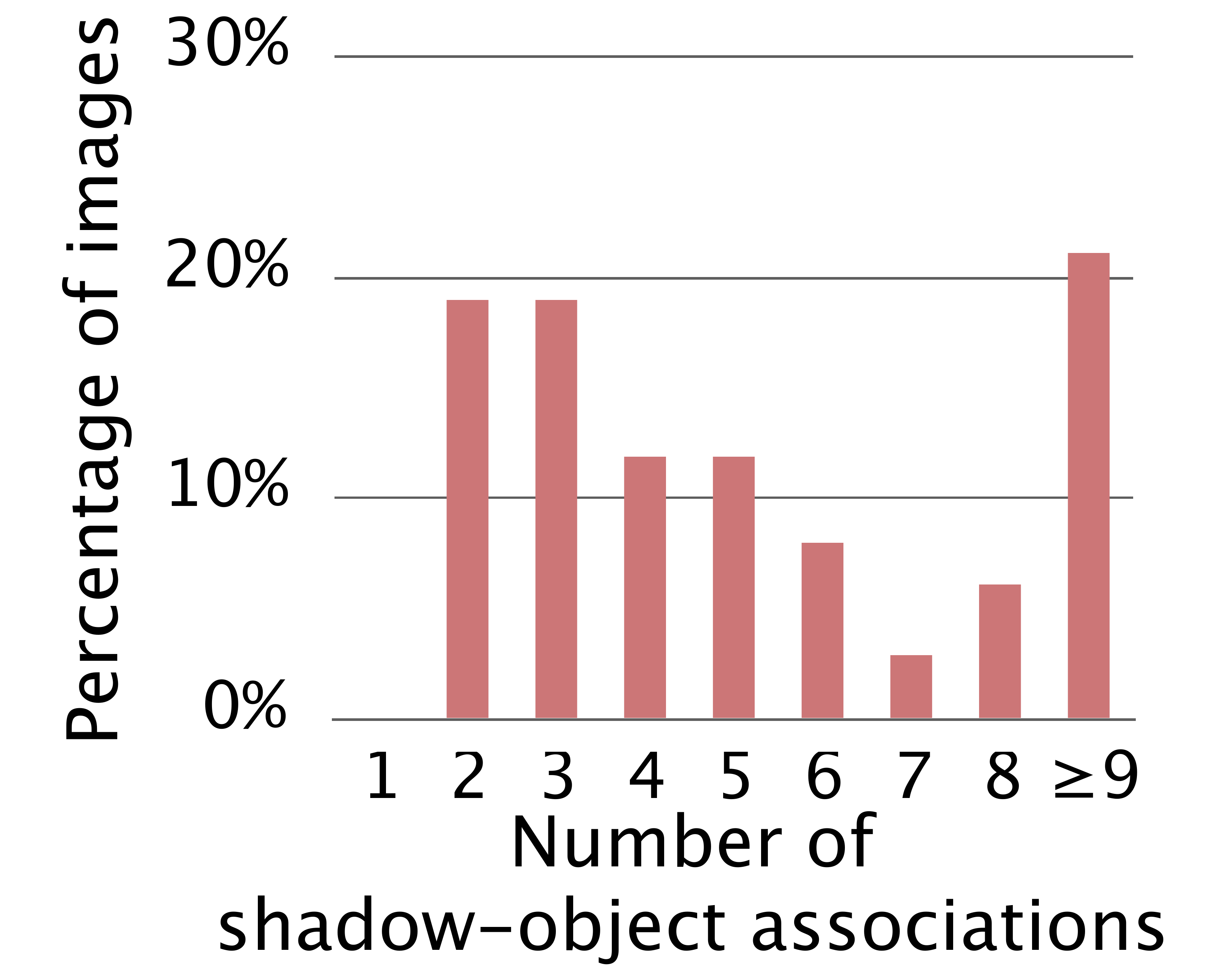}
\includegraphics[width=0.495\linewidth]{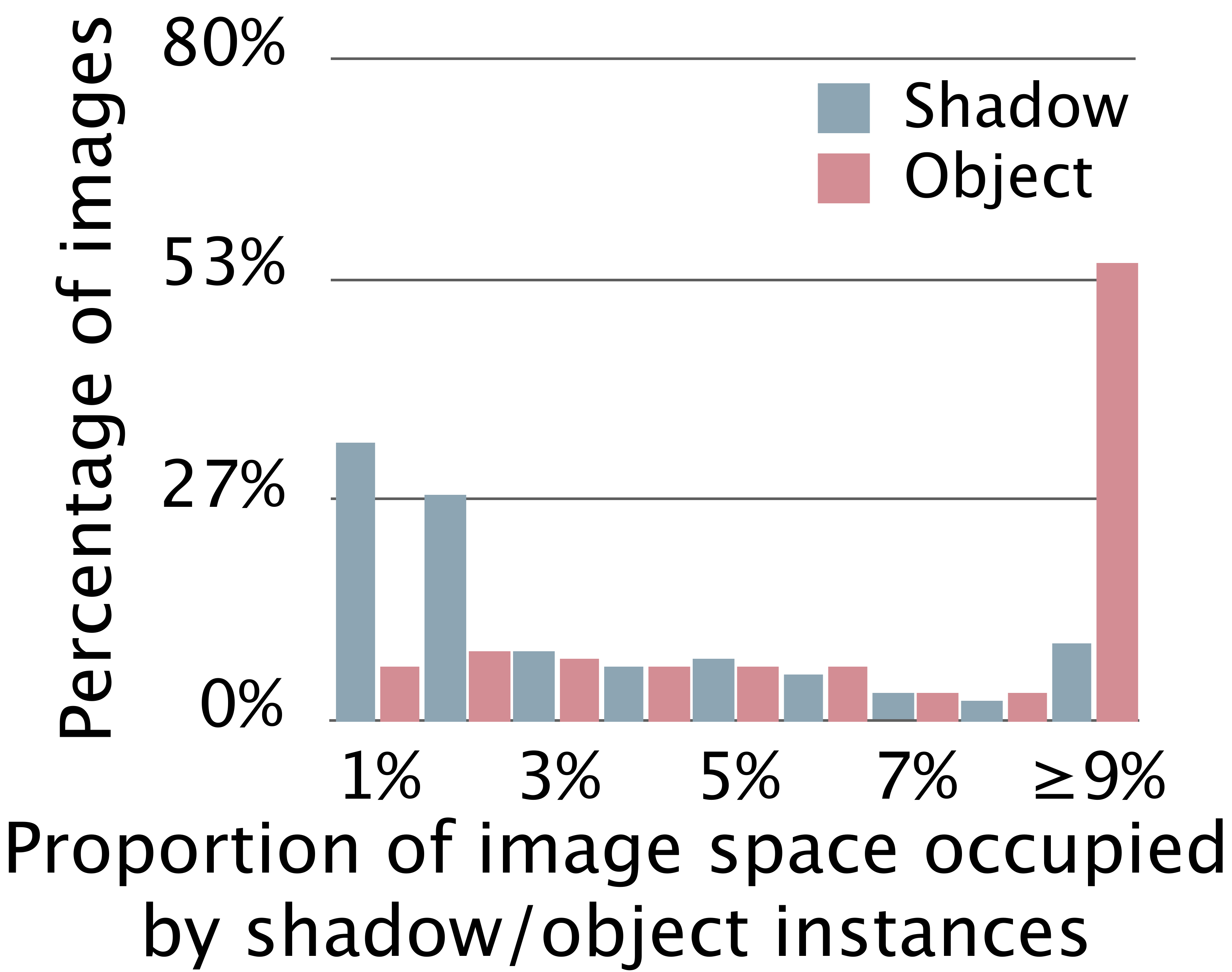}
\begin{minipage}[t]{\linewidth}
	\centering
	\centerline{\footnotesize (b) Statistical properties of the SOBA-challenge set.}
\end{minipage}
\caption{Statistical properties of the SOBA dataset.}
\label{img:dataset_analysis}
\vspace{-3mm}
\end{figure}

\subsection{SOBA (Shadow OBject Association) Dataset}

\wty{We prepare SOBA (Shadow OBject Association) dataset to support instance shadow detection, which contains three parts: SOBA training, SOBA testing, and SOBA challenge. 
%
We first build the SOBA-training and -testing sets from relatively simple cases by collecting images from the ADE20K~\cite{zhou2017scene,zhou2019semantic}, SBU~\cite{hou2021large, vicente2016noisy, vicente2016large}, ISTD~\cite{wang2018stacked}, and Microsoft COCO~\cite{lin2014microsoft} datasets, and also from the Internet using keyword search with shadow plus animal, people, car, athletic meeting, zoo, street, etc.
%
Then, we coarsely label the images to produce shadow instance masks and shadow-object association masks, and refine them using 
the Apple Pencil software; see Figures~\ref{img:1} (b) \& (d).
Next, we obtain object instance masks (see Figure~\ref{img:1} (c)) by subtracting each shadow instance mask from the associated shadow-object association mask.
Overall, there are 1,000 images with 3,623 pairs of shadow-object instances, and we randomly split the images into a training set (840 images, 2,999 pairs) and a testing set (160 images, 624 pairs); see Figure~\ref{img:dataset_SOBA} (a) \& (b) for some examples. 

We show some statistical properties of the SOBA-training and -testing sets in Figure~\ref{img:dataset_analysis} (a).
From the left histogram, we can see that it has a diverse number of shadow-object pairs per image, around 3.62 pairs per image on average.}
%
%
On the other hand, the right histogram reveals the proportion of image space (horizontal axis) occupied, respectively, by shadow and object instances in the dataset images.
From the plot, we can see that most shadows and objects occupy relatively small areas in the whole images.

\begin{figure*}[!t]
	\includegraphics[width=0.99\linewidth]{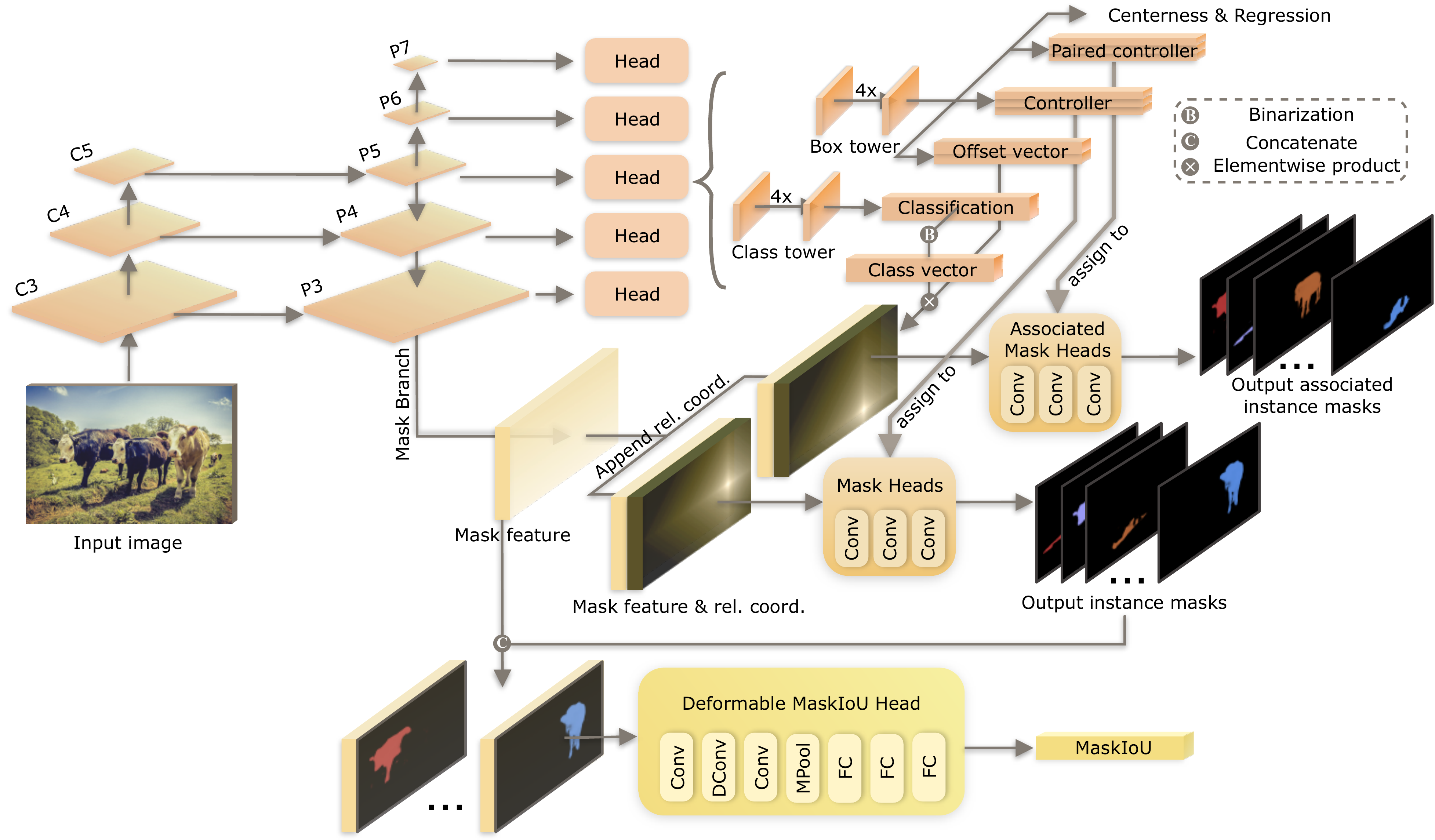}
	\vspace*{-1mm}
	\caption{The schematic illustration of our single-stage instance shadow detection network (SSISv2). 
		The mask feature and outputs of the {\em box tower\/} and {\em class tower\/} are used to formulate the bidirectional relation learning module; see Figure~\ref{fig:BRL}.
		The mask feature and output instance masks are sent to the {\em deformable MaskIoU head\/} for mask refinement; see Section~\ref{sec:maskiou}.
		Note that each head has its own box head and class head, and the filter parameters among these heads are shared.}
	\label{fig:arc}
	\vspace*{-1mm}
\end{figure*}

\wty{To evaluate the detection performance in complex scenarios, we further collected 100 images with challenging shadow-object pairs from the Internet using keyword search with {\em crowd\/} plus people, animals, cars, street, pasture, grassland, and beach. 
Then, we picked images of scenes with multiple various-shape shadow-object associations, large occlusion between the objects, between the shadows, or between both the objects and shadows, and long shadows that usually appear at sunset.}
Figure~\ref{img:dataset_SOBA} (c) shows some of these images.
%
%
%
Also, we annotated the images using similar steps as mentioned earlier. 
%
%
We name this dataset SOBA challenge, which includes 670 pairs of shadow-object instances, and the whole SOBA-challenge dataset is used only for testing.

Figure~\ref{img:dataset_analysis} (b) shows SOBA challenge's statistical properties.
From the left histogram, we can see that this dataset has $\sim$6.70 pairs per image on average (vs. 3.62 for SOBA training \& testing) and more than 20\% of the images have nine or more shadow-object pairs per image.
The right histogram also shows that this dataset contains more objects that occupy large areas in the images.

\wty{Overall, the whole SOBA dataset has 1,100 images with 4,293 pairs of annotated shadow-object associations.}
 


\subsection{SOAP (Shadow-Object Average Precision) Metric}

Existing metrics evaluate instance segmentation results by looking at object instances individually.
Our problem involves multiple types of instances: shadows, objects, and their associations.
Hence, we formulate a new metric called the \emph{Shadow-Object Average Precision} (SOAP) by adopting the same formulation as the traditional average precision (AP) with the intersection over union (IoU) but further considering a sample as true positive (an output shadow-object association), if it satisfies the following three conditions:
\begin{itemize}
	%
	\item[(i)]
	the IoU between the predicted shadow instance and ground-truth shadow instance is no less than $\tau$;
	%
	\item[(ii)]
	the IoU between the predicted object instance and ground-truth object instance is no less than $\tau$; and
	%
	\item[(iii)]
	the IoU between the predicted and ground-truth shadow-object associations is no less than $\tau$. 
\end{itemize}

We follow~\cite{lin2014microsoft} to report the evaluation results by setting $\tau$ as 0.5 ($\text{SOAP}_{50}$) or 0.75 ($\text{SOAP}_{75}$), and report also the average over multiple $\tau$ [0.5:0.05:0.95] ($\text{SOAP}$).
Also, since we obtain the bounding boxes and masks of the shadow instances, object instances, and shadow-object associations, we report $\text{SOAP}_{50}$, $\text{SOAP}_{75}$, and $\text{SOAP}$ for both bounding boxes and masks.
\emph{The dataset and evaluation metric are available for download at
\url{https://github.com/stevewongv/InstanceShadowDetection}.}



\section{Methodology}
\label{sec:method}

\subsection{Overall Network Architecture}


Figure~\ref{fig:arc} overviews our network architecture.
%
Given the input image, we leverage a convolutional neural network to extract feature maps in varying solutions and employ a feature pyramid network~\cite{Lin2017fpn} with multiple feature levels ($P3$ to $P7$).
%
Then, we adopt multiple heads at different levels: a class tower with four convolutional layers to predict the classification scores and a box tower with another four convolutional layers for other predictions.
In summary, we obtain the following predictions for each head:
\begin{itemize}
\item[(i)]
\textbf{classification} scores, which indicate the categories of shadow, object, and background;
%
\item[(ii)]
\textbf{offset vector}, which are image-space vectors from the centers of shadow instances to the centers of the corresponding object instance, and vice versa;
%
\item[(iii)]
\textbf{controller} and \textbf{paired controller}, each learning a set of filter parameters in the mask head to predict the masks for shadow instance and object instance, respectively. Note that each instance has its individual filter parameters to predict a mask; see~\cite{tian2020conditional} for details. In our framework, if the controller generates filter parameters for a shadow instance, the paired controller will generate filter parameters for the corresponding object instance, and vice versa; and
%
\item[(iv)]
\textbf{regression} and \textbf{centerness}: regression predicts the bounding box of each shadow and object instance, whereas centerness regularizes the prediction by reducing the number of low-quality predicted bounding boxes far from the center of a target shadow/object; see~\cite{tian2020fcos} for details.
\end{itemize}

Next, we formulate a mask branch, which takes the feature map at $P3$ as input and generates the mask feature.
\wty{For each predicted shadow/object instance, we duplicate and concatenate the mask feature with two relative coordinate (Rel. Coord.) maps: 
one indicates the center of the object/shadow instance,
whereas the other is obtained by first multiplying the offset vector with a class vector then adding the results with the coordinates to represent the center of the corresponding shadow/object instance.}
Note that the class vector is generated from the classification score, where $-1$ ($+1$) indicates the direction from object to shadow (from shadow to object) and the relative coordinate map is computed from the predicted locations of shadow/object instances.
Further, we use the learned filter parameters from the controller and paired controller to perform convolutional operations on the concatenated feature mask and relative coordinate maps and predict the masks for the shadow/object instances and the paired object/shadow instances.
Finally, we concatenate the predicted masks for the instances and the mask feature and design a deformable MaskIoU head to refine the predicted masks by adopting a MaskIoU loss function.

In the following, we will elaborate on how to learn the relation between shadow and object instances (Section~\ref{sec:brl}) and formulate the deformable MaskIoU head (Section~\ref{sec:maskiou}), and then present the training and testing strategies, including the shadow-aware copy-and-paste augmentation and loss functions (Section~\ref{sec:train_test}).

\begin{figure*}[!t]
	\includegraphics[width=\linewidth]{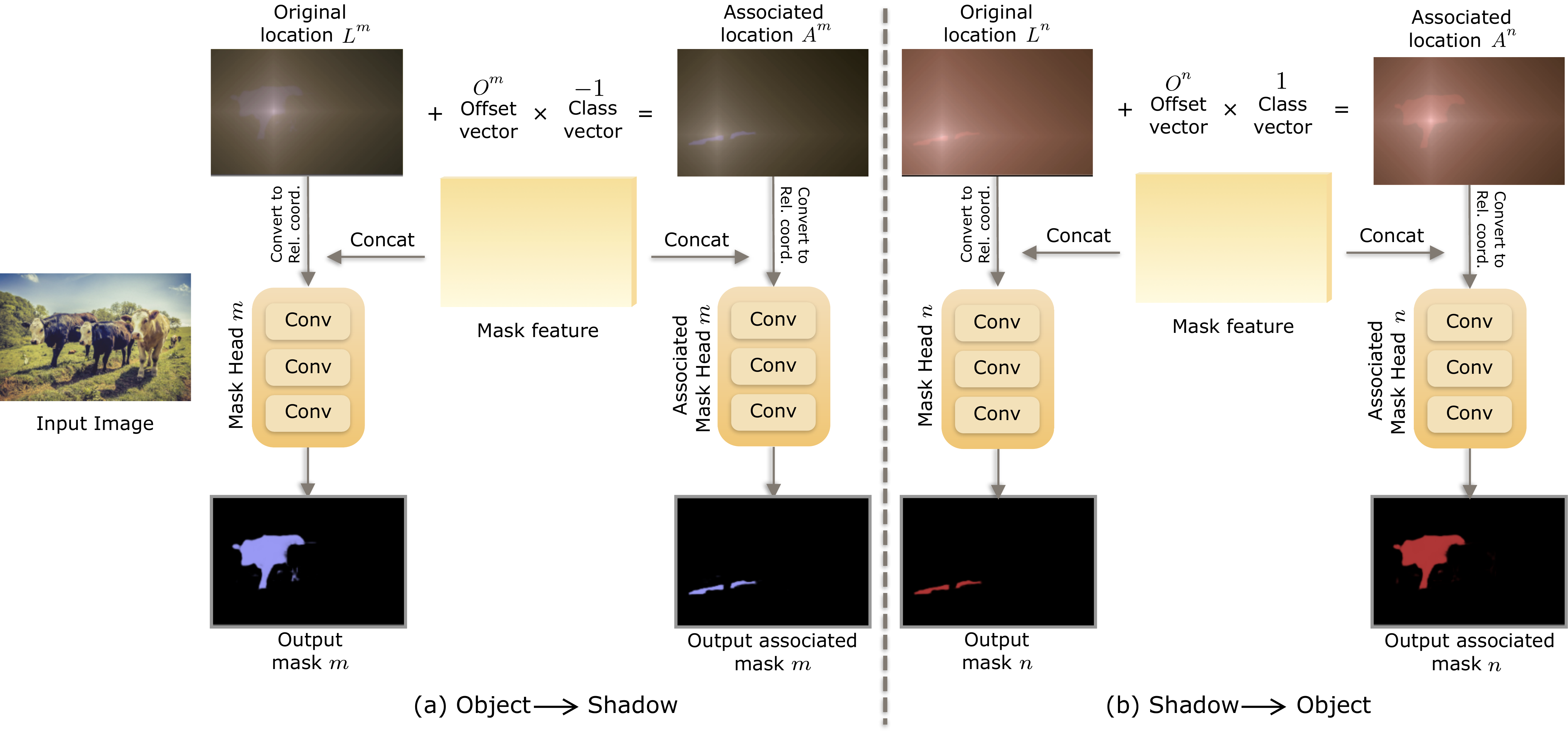}
	\caption{The schematic illustration of the bidirectional relation learning module in our network. The left part (Object $\rightarrow$ Shadow) shows how to find the associated shadow instance from the location of the paired object instance, whereas the right part (Shadow $\rightarrow$ Object) shows how to find the associated object instance from the location of the paired shadow instance.} 
	\label{fig:BRL}
	\vspace*{-3mm}
\end{figure*}

\subsection{Bidirectional Relation Learning}
\label{sec:brl}

%
Figure~\ref{fig:BRL} shows the detailed structure of our proposed bidirectional relation learning module.
Figure~\ref{fig:BRL} (a) illustrates how to learn the paired shadow instance from the object instance, whereas Figure~\ref{fig:BRL} (b) illustrates this strategy in the opposite direction.
As shown in the top left corner, after obtaining the original location $L^m$ of the $m$-th object instance, we append the location with the mask feature and adopt the $m$-th mask head to predict the segmentation mask of this instance.
Note that the filter parameters in the mask head are produced from the controller and the filter parameters vary in different mask heads; see ``Controller'' in Figure~\ref{fig:arc}.

Then, we compute the associated location $A^m$ to mark the center of the paired shadow instance by using the learned offset vector $O^m$ and class vector $-1$:
\begin{equation}
A^m \ = \ L^m \ + \ O^m \ \times \ -1 \ ,
\end{equation}
where the offset vector is learned from the box tower and it represents the distance between the center of the object instance and the center of the paired shadow instance; 
the class vector is generated from the classification score and we adopt $-1$ to represent the direction from object to shadow and $1$ to represent the direction from shadow to object.
Next, we concatenate the associated location $A^m$ and mask feature, and use the $m$-th associated mask head to generate the mask for the shadow instance, and the filter parameters of the associated mask head are learned from the paired controller automatically, as shown in Figure~\ref{fig:arc}.

Similarly, taking the original location $L^n$ of the $n$-th shadow instance as the input, we compute the associated location $A^n$ of the paired object instance by
\begin{equation}
A^n \ = \ L^n \ + \ O^n \ \times \ 1 \ ,
\end{equation}
where $O^n$ denotes the $n$-th offset vector and $1$ denotes the learning direction from shadow to object. 
Also, we adopt the mask head and the associated mask head to generate the segmentation masks for the paired shadow and object instances; see Figure~\ref{fig:BRL} (right).

Note that the location maps ($L^m, A^m, L^n, A^n$) shown in Figure~\ref{fig:BRL} are the visualization results of the learned locations, demonstrating that our network can successfully learn the locations for the shadow and object pairs.

\begin{figure}[tp]     
\centering
\includegraphics[width=0.9\linewidth]{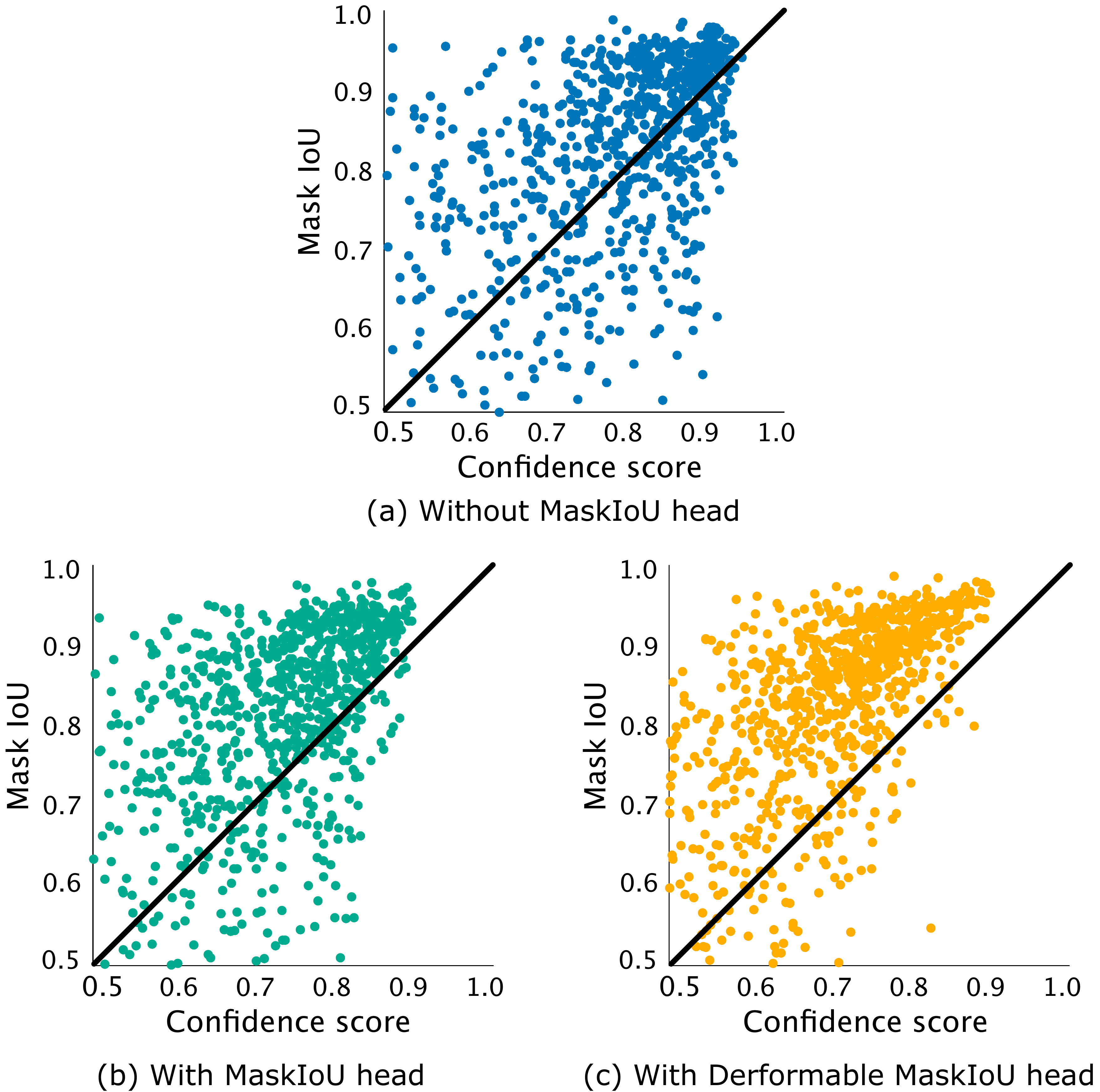}

\caption{Confidence scores vs. mask IoUs before and after applying the MaskIoU head or Deformable MaskIoU Head. Each point in the figure denotes a predicted instance mask. The thick lines in the plots indicate Confidence score equals mask IoU. As shown in (c), with the Deformable MaskIoU head, we can dramatically avoid more masks with high confidence scores but low IoUs.}
\label{img:dmaskiou_analysis}
\vspace{-3mm}
\end{figure}

%
%

\subsection{Deformable MaskIoU Head}
\label{sec:maskiou}


\wty{As shown in Figure~\ref{img:dmaskiou_analysis} (a), the original model tends to predict masks with high confidence scores but low IoUs.
These low-quality masks degrade the detection performance,
since the confidence score predicted from the classification task (``classification'' in Figure~\ref{fig:arc}) fails to consider the mask information. 
To further refine the predicted masks of shadow/object instances, we formulate the deformable MaskIoU head to regress the intersection over union (IoU) between the predicted masks and the associated ground-truth masks.}
As shown in Figure~\ref{fig:arc} (bottom), given the mask feature and each predicted mask as input, we perform a $1 \times 1$ convolution to reduce the feature channel and a deformable convolution layer~\cite{zhu2019defor} to focus the learning on the instance's specific region, followed by a convolution layer and an adaptive max-pooling layer to reshape the feature map to $64 \times 64$.
Lastly, we leverage three fully connected layers to predict a single mask IoU per instance.

\wty{Unlike the MaskIoU head in~\cite{huang2019msrcnn}, which is designed only for RoI-based methods and takes the RoI feature of size $14\times14$ as input, 
we design a deformable MaskIoU head that takes the whole mask feature with instance mask as input and automatically learns the discriminative feature for each instance mask, since our based method~\cite{tian2020conditional} employs the whole mask feature with conditional convolution to eliminate the RoI operations. }
%
Figure~\ref{img:dmaskiou_analysis} shows the predicted instance masks with confidence scores and associated mask IoUs before and after using the MaskIoU head.
%
\wty{As shown on the bottom right of the figure, after using the deformable MaskIoU head, we can dramatically avoid more masks with high confidences but low IoUs, showing that our deformable MaskIoU head can successfully filter out instances of low quality.
Please see Section~\ref{Sec:ablation} for related quantitative comparison results.}

\subsection{Training and Testing Strategies}
\label{sec:train_test}

\begin{figure}[tp]     
\centering
\includegraphics[width=0.98\linewidth]{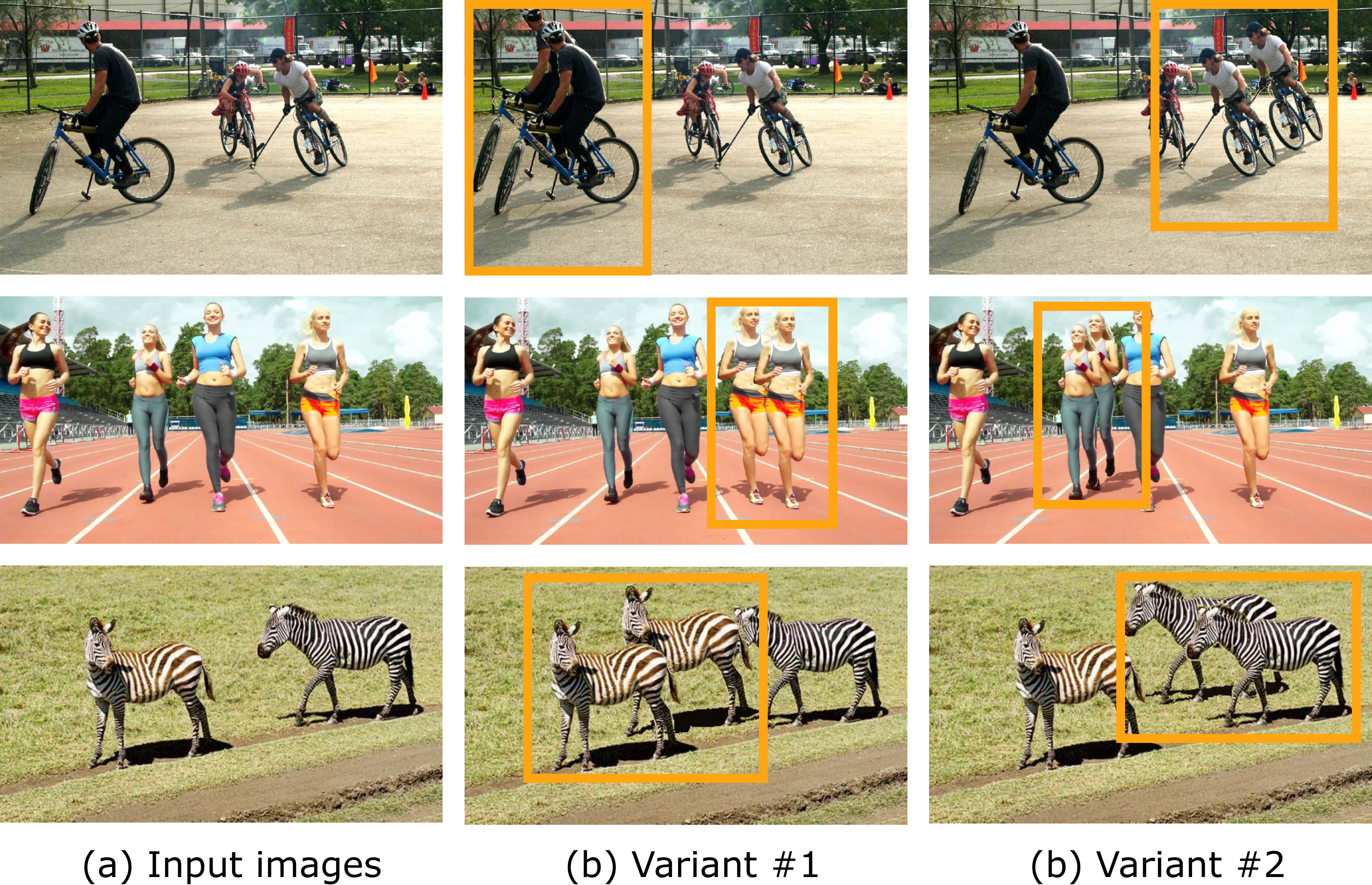}
\caption{\wty{Shadow-aware copy-and-paste augmentation. (b) \& (c) show example copying-and-pasting results on 
different objects.}}
\label{img:data_augm}
\vspace{-3mm}
\end{figure}

\subsubsection{Shadow-aware Copy-and-Paste Augmentation}~\label{sec:copypaste}

\vspace*{-3.5mm}
To enhance the network's robustness, especially for handling challenge cases,~\eg, occlusion between object and shadow instances, we design a shadow-aware copy-and-paste augmentation strategy to enrich the training data.
As shown in Figure~\ref{img:data_augm}, we randomly select a shadow-object association in the input image, copy the object instance with its associated shadow instance, then paste them to the surroundings.
\wty{Specifically, we shift the mask by a random value in range [-$\frac{2}{3}$W, $\frac{2}{3}$W] on the X axis and a random value in range (0, $\frac{2}{3}$H] on the Y axis, where W and  H are the width and height of the shifted object.}
Importantly, the augmentation should consider {\em object layering\/}.
That is, we should put the pasted shadow-object association behind existing object instances but above their original shadow instances and scene background for plausible occlusions among the objects.
%
Further, we propose to relight the scene background in the shadow region cast by the copied object.
The relighted shadow region $R$ is computed by
\begin{equation}
 R = \frac{mean(S)}{mean(T)} \cdot T \ ,
\end{equation}
where $T$ is the original color of the relighted shadow region and $S$ is the color of the shadow region where is copied.




%


\subsubsection{Loss Function}
%


We define the overall loss $\mathcal{L}_{\text{all}}$ for training our SISSv2 network as a sum of detection loss $\mathcal{L}_{\text{D}}$, mask loss $\mathcal{L}_{\text{M}}$, and boundary loss $\mathcal{L}_{\text{B}}$:
\begin{equation}
 \mathcal{L}_{\text{all}} = \mathcal{L}_{\text{D}} + \mathcal{L}_{\text{M}} + \mathcal{L}_{\text{B}} \ .
\end{equation}


\vspace*{-2mm}
\noindent
{\bf Detection loss:} \ 
\begin{equation}
 \mathcal{L}_{\text{D}} = \mathcal{L}_{\text{cls}} + \mathcal{L}_{\text{center}} + \mathcal{L}_{\text{box}} + \mathcal{L}_{\text{offset}} ,
\end{equation}
where $\mathcal{L}_{\text{cls}}$ is the classification loss, $\mathcal{L}_{\text{center}}$ is the centerness loss, and $\mathcal{L}_{\text{box}}$ is the box regression loss, which follows the losses in~\cite{tian2020fcos}.
The offset loss $\mathcal{L}_{\text{offset}}$ takes the form of the smooth $\mathcal{L}_{1}$ loss~\cite{girshick2015fast} for optimizing the offset vectors:
\begin{equation}
  \mathcal{L}_{\text{offset}}\left( u, v\right)   \ = \ \sum_{I \in \{x,y\}}
  \left\{
    \begin{array}{ll}
	0.5\left( u_i-v_i\right)^{2} \ , & \text { if }\left|u_i-v_i\right|<1 ; \ \\
	\left|u_i-v_i\right|-0.5 \ , & \text { otherwise } , 
	\end{array}\right. 
\end{equation}
where $u_i$ is resulted from the element-wise multiplication of the predicted offset vector and class vector:
\begin{equation}
u_i \ = \ O_i \ \times \ C_i \ ,
\end{equation}
and $v_i$ denotes the ground-truth offset vector:
\begin{equation}
v_i \ = \ \tilde{L}_i  \ - \ L_i \ ,
\end{equation}
where $\tilde{L}_i$ and $L_i$ are the ground-truth and predicted location of the paired object/shadow instance, respectively.

\vspace*{2mm}  
\noindent
{\bf Mask loss:} \ 
\begin{equation}
 \mathcal{L}_{\text{M}} = \mathcal{L}_{\text{mask}} + \mathcal{L}_{\text{mask}}^{\text{associated}} + \mathcal{L}_{\text{maskiou}} \ , 
\end{equation}
where we adopt dice loss~\cite{milletari2016v} to compute the losses of the output instance masks $\mathcal{L}_{\text{mask}}$ and produce the associated instance masks $\mathcal{L}_{\text{mask}}^{\text{associated}}$; see Figure~\ref{fig:arc} for the predictions. 
The MaskIoU loss $\mathcal{L}_{\text{maskiou}}$ is defined as
\begin{equation}
\mathcal{L}_{\text{maskiou}} = \frac{1}{N}\sum_{i=1}^{N}(I_i - \tilde{I}_i)^2 \ ,
\end{equation}
where $I_i$ and $\tilde{I}_i$ are the predicted and ground-truth mask IoU, respectively; and $N$ is the number of the predicted instances. 

 \begin{figure}[tp]
 	\centering
 	\begin{minipage}[t]{ \linewidth}
 		\includegraphics[width=\linewidth]{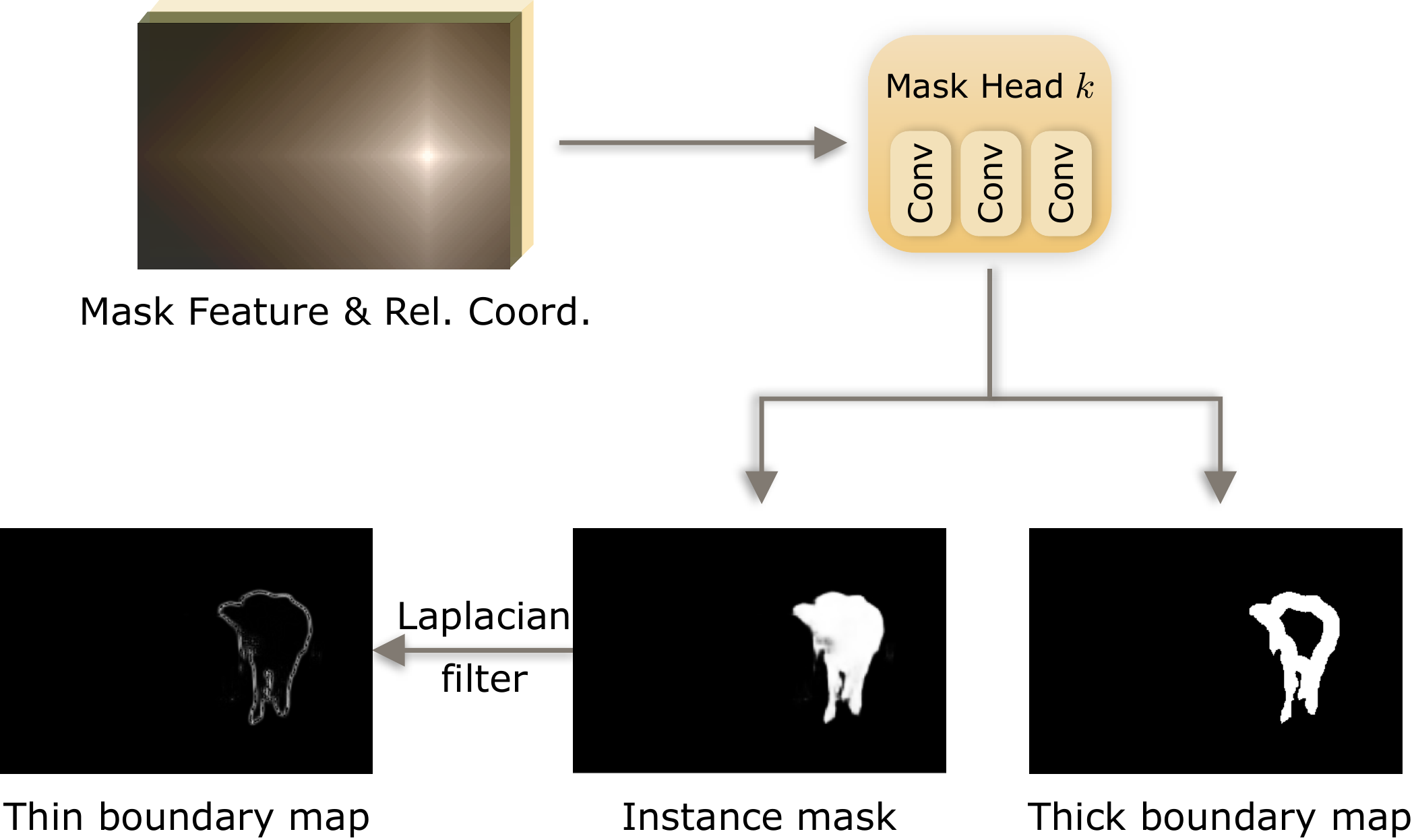}
 	\end{minipage}
 	\caption{\wty{
 	The mask head (top right) simultaneously predicts {\em a thick boundary map\/} and an instance mask.
 	We then pass the instance mask to the Laplacian filter to produce the {\em thin boundary map\/}.}}
 	\label{fig:boundary_generation}
 	\vspace*{-3mm}
 \end{figure}

 \begin{table*}[t]
	\centering
	\caption{Comparison with the previous state-of-the-art methods for instance shadow detection on the SOBA-testing set.
	\wty{Note that the results of LISA and SSIS are slightly different from the results reported in the conference versions~\cite{wang2020instance,wang2021single}, because this work replaces a simple polygon representation (employed in the previous works) with a more precise representation,~\ie, RLE, for the labeled masks.}}
	\vspace*{-1mm}
	\label{table:state_of_the_art}
	\resizebox{\linewidth}{!}{%
		\begin{tabular}{l|cc|cccc} 
			\toprule
			Network & $SOAP_{segm}$ & $SOAP_{bbox}$ & Association $AP_{segm}$ & Association $AP_{bbox}$  & Instance $AP_{segm}$ & Instance $AP_{bbox}$ \\ 
			\midrule
			LISA~\cite{wang2020instance} & 23.5 & 21.9 & 40.9 & 48.4 & 39.2 & 37.6 \\
			
			SSIS~\cite{wang2021single} & 30.2   & 27.1  & 52.2  & 59.6  & 43.4   & 41.3 \\
			\textbf{SSISv2} &  \textbf{35.3} &  \textbf{29.0}&  \textbf{59.2} & \textbf{63.1}  &  \textbf{50.2} &  \textbf{44.4}  \\
			\bottomrule
	\end{tabular}}
	
\end{table*}

\begin{table*}
	\centering
	\caption{Comparison with the previous state-of-the-art methods for instance shadow detection on the SOBA-challenge set.}
	\label{table:challenge}
	\vspace*{-1mm}
	\resizebox{\linewidth}{!}{%
		\begin{tabular}{l|cc|cccc} 
			\toprule
			Network & $SOAP_{segm}$ & $SOAP_{bbox}$ & Association $AP_{segm}$ & Association $AP_{bbox}$  & Instance $AP_{segm}$ & Instance $AP_{bbox}$ \\ 
			\midrule
			LISA~\cite{wang2020instance} & 10.4 & 10.1 & 20.7 & 25.8 & 23.8 & 24.3 \\
			
			SSIS~\cite{wang2021single} & 12.7   & 12.8  & 28.4  & 32.6  & 25.6   & 26.2 \\
			\textbf{SSISv2} & \textbf{17.7}   & \textbf{15.1}  & \textbf{34.6}  & \textbf{37.3} & \textbf{31.0} & \textbf{28.4}  \\
			\bottomrule
	\end{tabular}}
\vspace*{-2mm}
\end{table*}

\vspace*{2mm}
\noindent
{\bf Boundary loss.} \
\wty{Different from existing boundary losses, we formulate two types boundary maps with different thicknesses to improve the boundary accuracy of the instance masks.
One is a thick boundary map for focusing on the boundary structures, and the other is a thin boundary map for focusing on the boundary details.
%
%
%
As shown in Figure~\ref{fig:boundary_generation}, we predict the thick boundary map from the mask head directly and generate the thin boundary map by applying a Laplacian filter on the predicted instance mask.}
On the other hand, we extract the boundary map from the ground-truth image and apply the Laplacian filter to the boundary map to formulate a supervision on the predicted thin boundary map; then, we apply the Euclidean distance transform~\cite{rosenfeld1968distance} to the boundary map from the ground truth to formulate a supervision on the predicted thick boundary map.
%
%
%
The overall boundary loss is a summation of the output instance masks $\mathcal{L}_{\text{boundary}}$ and the output associated instance masks $\mathcal{L}_{\text{boundary}}^{\text{associated}}$:
%
\begin{equation}
\mathcal{L}_{\text{B}} = \mathcal{L}_{\text{boundary}} + \mathcal{L}_{\text{boundary}}^{\text{associated}} \ ,
\end{equation}
\begin{equation}
\mathcal{L}_{\text{boundary}} = \beta \frac{ \arrowvert|l(\tilde{m})| - |l(m)|\arrowvert}{ | l(\tilde{m})| } + dice( \frac{d(\tilde{m})}{max(d(\tilde{m}))} < 0.5,b) \ ,
\end{equation}
where $\tilde{m}$ is ground-truth instance mask; $m$ is predicted instance mask;
$l(x)$ is Laplacian filter, whose kernel size is five; \wty{weight $\beta$ is set as five to balance the loss values;}
$d$ 
computes a distance field, in which each pixel stores the distance to the nearest boundary pixel; $max(d(\tilde{m}))$ is the maximum distance for normalization;
$dice$ is dice loss; and $b$ is the predicted thick boundary map.

Figure~\ref{fig:boundary} shows example results produced by our method with and without the boundary loss; by adopting the boundary loss in training, we can improve the accuracy of the predicted instance masks; \wty{please see Section~\ref{Sec:ablation} for the quantitative comparison}.
%
%
%
Both offset loss $\mathcal{L}_{\text{offset}}$ and mask loss of the associated instance mask $\mathcal{L}_{\text{mask}}^{\text{associated}}$ propagate the gradient to offset vectors, helping to optimize the network during the training.
Also, we do not use $\mathcal{L}_{\text{maskiou}}$ in the first $5,000$ training iterations and thin boundary loss in the first $10,000$ training iterations, as the predicted instance masks have low quality at the beginning of the training process.

\subsubsection{Training Parameters} 
We train our network by adopting the strategies of CondInst~\cite{tian2020conditional} and AdelaiDet~\cite{Tian2020Adet}.
First, we adopt the weights of ResNeXt-101-BiFPN~\cite{xie2017aggregated,tan2020efficientdet} trained on ImageNet~\cite{deng2009imagenet} to initialize the backbone network parameters, set the mini-batch size as two, and optimize our network on one NVidia RTX 3090 GPU. 
Second, we set the base learning rate as $0.001$, adopt a warm-up~\cite{goyal2017accurate} strategy to linearly increase the learning rate from $0.0001$ to $0.001$ in the first $1,00$ iterations, reduce the learning rate to $0.0001$ after $40,000$ iterations, and stop the learning after $45,000$ iterations.
Third, we re-scale the input images, such that the longer side is smaller than $1,333$ and the shorter side was smaller than $640$, without changing the image aspect ratio.
Lastly, we apply random horizontal flip to the input images as data augmentation.

\subsubsection{Inference} 
\label{sec:inference}
In testing, the mask heads in our network produce the masks for the shadow and object instances, while the associated mask heads generate the masks for the paired object and shadow instances based on the learned offset vectors; see Figure~\ref{fig:BRL}.
With bidirectional relation learning, we can obtain two sets of predicted masks, for each pair of shadow and object instances.
%
If the main branch (left branch in Figure~\ref{fig:BRL} (a)\&(b)) produces the mask of its shadow instance, the associated branch (right branch in Figure~\ref{fig:BRL} (a)\&(b)) will generate the mask of its object instance, and vice versa.
%
%
Yet, the accuracy of mask predictions in the main branch is usually better than that of the associated branch, since the associated branch needs to learn both tasks of mask prediction and shadow-object relation, making its training more difficult.
Hence, we adopt the associated branch only to predict the paired relation of the shadow and object instances, and take the masks predicted from the main branch as the results.
Finally, we adopt mask non-maximum suppression (NMS) to refine the results.



\begin{figure*}[tp]
	 \ \ \ \ \ \    \includegraphics[width=0.98\linewidth]{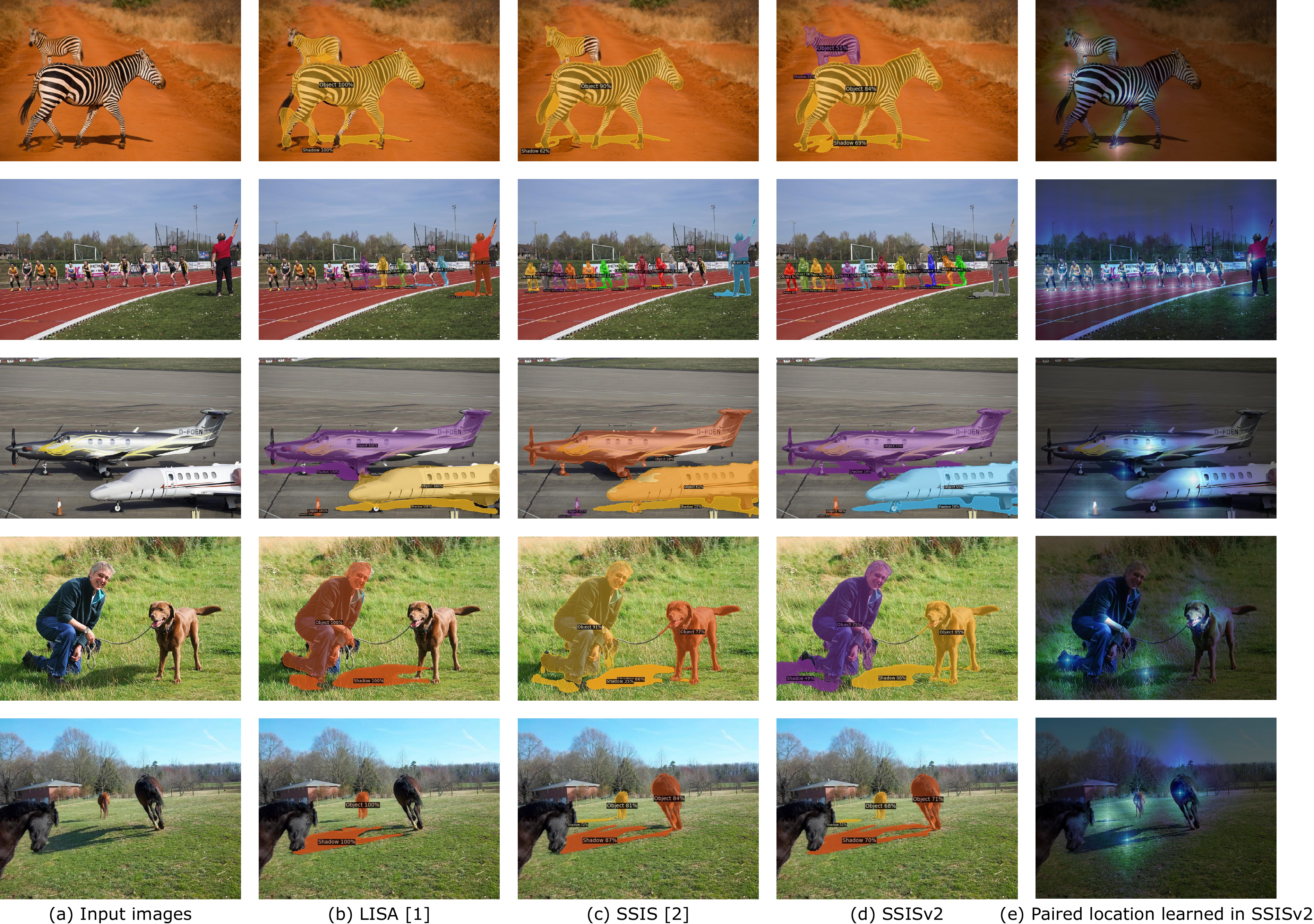}
	\vspace*{-6mm}
	\caption{Visual comparison between instance shadow detection results produced by various methods (b)-(d) on images (a) in the SOBA-testing set; (e) shows the learned locations for pairing shadow and object instances in our method.}
	\label{fig:visual_comparision}
	\vspace*{-1mm}
\end{figure*}

\begin{figure*}[tp]
	\centering
	\includegraphics[width=0.95\linewidth]{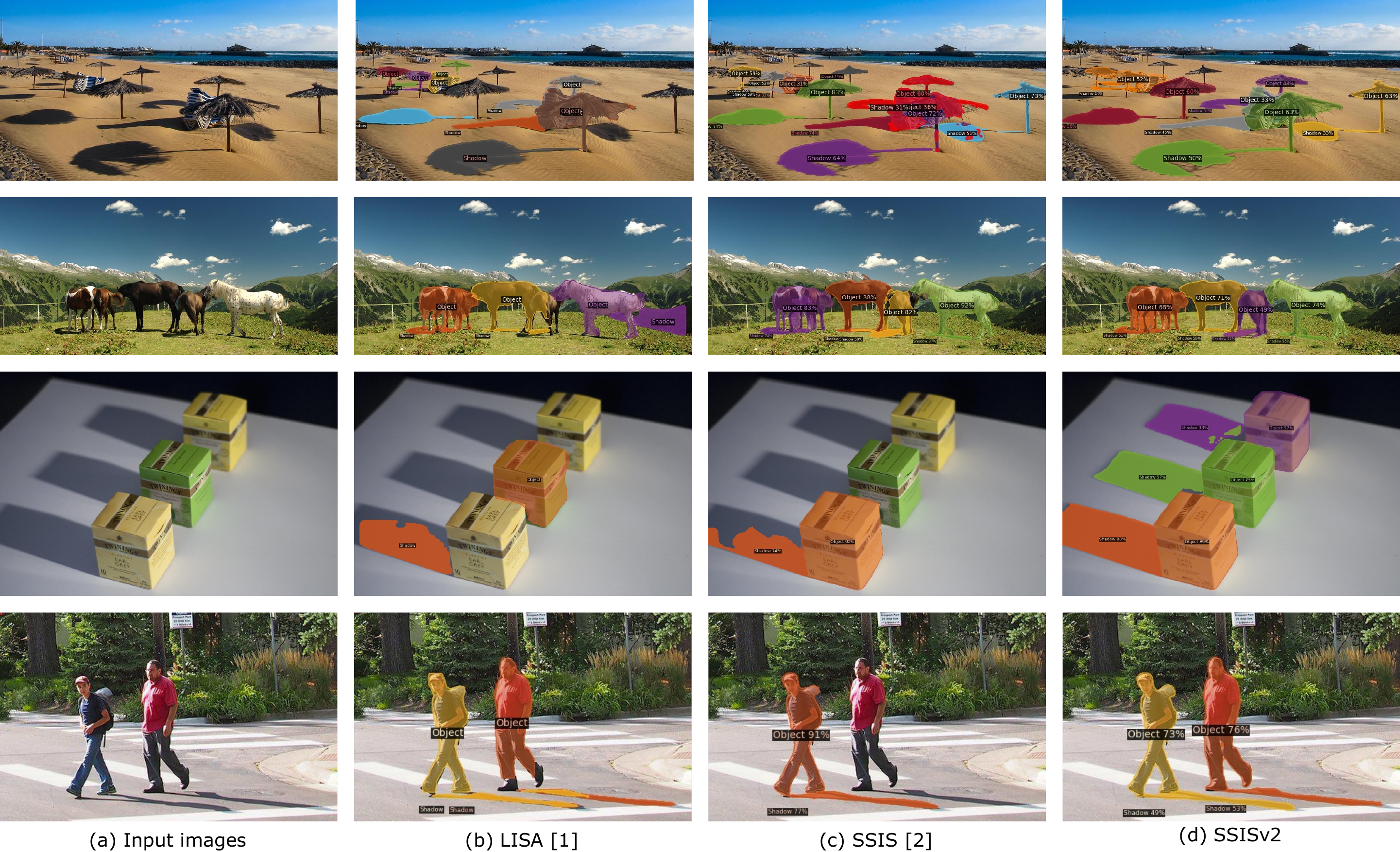}
	\vspace*{-2mm}
	\caption{Visual comparison between instance shadow detection results produced by various methods (b)-(d) on the SOBA-challenge set.}
	\label{fig:visual_comparision_SOBA_challenge}
	\vspace*{-2mm}
\end{figure*}

\section{Experimental Results}
\label{sec:experiments}

\begin{table*}[htp]
	\centering
	\caption{Component analysis on the SOBA-testing set; ``data augm'' denotes shadow-aware copy-and-paste (see Section~\ref{sec:copypaste}).}
	\vspace*{-1mm}
	\label{table:component_analysis}
\resizebox{\textwidth}{!}{%
\begin{tabular}{@{}cccccccccc@{}}
\toprule
 & \begin{tabular}[c]{@{}c@{}}+ deformable\\ maskIoU\end{tabular} & \begin{tabular}[c]{@{}c@{}}+ boundary\\ loss\end{tabular} & \begin{tabular}[c]{@{}c@{}}+ data \\ augm\end{tabular} & $SOAP_{segm}$ & $SOAP_{box}$ & \begin{tabular}[c]{@{}c@{}}Association\\ $AP_{segm}$\end{tabular} & \begin{tabular}[c]{@{}c@{}}Association\\ $AP_{box}$\end{tabular} & \begin{tabular}[c]{@{}c@{}}Instance\\ $AP_{segm}$\end{tabular} & \begin{tabular}[c]{@{}c@{}}Instance\\ $AP_{box}$\end{tabular} \\ \midrule
			basic        &  &  & & 28.1 & 26.8 & 49.8 & 56.6 & 41.5  & 40.8  \\
			+ offset     &  &  & & 29.1  & 25.4 & 51.1 & 57.6 & 41.8 & 38.9 \\
 \begin{tabular}[c]{@{}c@{}}+ class  (SSIS~\cite{wang2021single})\end{tabular} &  &  &  & 30.2 & 27.1 & 53.6 & 59.6 & 43.4 & 41.3 \\
 \midrule
 & \checkmark &  &  			& 32.0 & 27.2 & 53.6 & 58.8 & 45.9 & 41.1 \\
 &  & \checkmark &  			& 31.0 & 26.3 & 54.6 & 59.4 & 45.0 & 40.8 \\
 &  &  & \checkmark 			& 31.1 & 26.7 & 55.1 & 60.6 & 45.3 & 42.0 \\
 & \checkmark & \checkmark &  	& 33.3 & 27.5 & 56.0 & 60.7 & 46.4 & 42.0 \\
 & \checkmark &  & \checkmark 	& 33.5 & 27.8 & 56.8 & 62.2 & 47.8 & 42.8 \\
 &  & \checkmark & \checkmark 	& 32.1 & 27.7 & 56.4 & 61.7 & 46.3 & 42.8 \\
SSISv2 & \checkmark & \checkmark & \checkmark & \textbf{35.3} & \textbf{29.0} & \textbf{59.2} & \textbf{63.1} & \textbf{50.2} & \textbf{44.4} \\ \bottomrule
\end{tabular}%
}
\vspace*{-2mm}
\end{table*}

\subsection{Comparison with State-of-the-art Methods}
We compare our SSISv2 with the instance shadow detection methods in our conference versions, LISA~\cite{wang2020instance} and SSIS~\cite{wang2021single}.
LISA is a two-stage detector that takes light direction as guidance and adopts a post-processing strategy to pair up the predicted shadow/object instances with the shadow-object associations, whereas SSIS is a single-stage fully convolutional detector that directly predicts shadow instances, object instances, and their associations. 
\wty{Our SSISv2 further formulates the deformable MaskIoU head, the shadow-aware copy-and-paste data augmentation strategy, and the boundary loss to improve the performance over SSIS~\cite{wang2021single}.}
%

Table~\ref{table:state_of_the_art} reports the comparison results on the SOBA-testing set.
We can see that SSISv2 clearly outperforms previous state-of-the-art methods, LISA~\cite{wang2020instance} and SSIS~\cite{wang2021single}, for all the evaluation metrics,
%
where the improvements on $SOAP_{segm}$ and $SOAP_{bbox}$ are $50.2\%$ / $32.4\%$ over LISA and $16.9\%$ / $7.0\%$ over SSIS, respectively, showing the superiority of SSISv2.
Further, we compare the methods on the SOBA-challenge set and report the results in Table~\ref{table:challenge}; SSISv2 also achieves the best results in terms of all the evaluation metrics in the challenge scenarios.

\begin{figure*}[tp]
	\centering
	\includegraphics[width=0.9\linewidth]{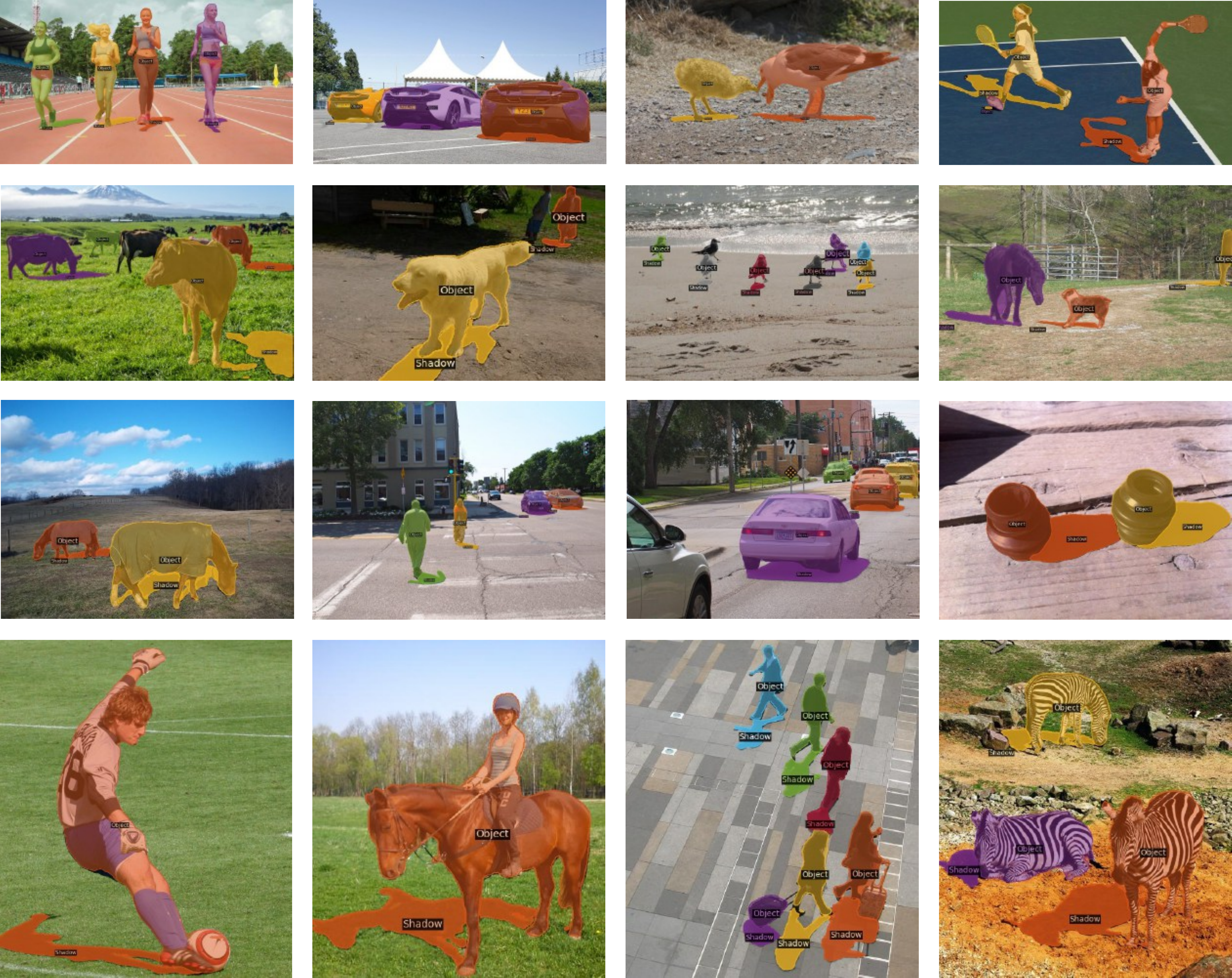}
	\caption{More instance shadow detection results produced by our SSISv2 over a wide variety of objects and shadows. The top two rows are from the SOBA-testing set while the others are from the SOBA-challenge set.}
	\label{fig:more_res}
\end{figure*}

Next, we provide visual comparison results in Figure~\ref{fig:visual_comparision}, where (a) shows the input images;
(b), (c), and (d) show the results produced by LISA~\cite{wang2020instance}, SSIS~\cite{wang2021single}, and SSISv2, respectively, and (e) shows the paired locations learned by SSISv2 to indicate the paired shadow and object instances.
From the results, we can see that
(i) SSISv2 can discover more shadow-object association pairs, as shown in the first two rows;
(ii) SSISv2 can produce more accurate masks for shadow and object instances, as shown in the third row;
(iii) SSISv2 can successfully pair up the object and shadow instances, but previous methods may fail; see the last two rows; and
(iv) SSISv2 can learn the locations of shadow-object pairs through the directional relation learning module, as shown in (e).
Figure~\ref{fig:visual_comparision_SOBA_challenge} shows the visual comparison results on the SOBA-challenging set, where SSISv2 better pairs up the shadow and object instances and produces more accurate instance masks than the results produced by the previous methods.
Please see Figure~\ref{fig:more_res} for more instance shadow detection results produced by SSISv2 on various types of objects and shadows.
\emph{Our code, trained models, and the results are released at \url{https://github.com/stevewongv/SSIS}}.

\subsection{Evaluation on the Network Design} \label{Sec:ablation}

\noindent
{\bf Component analysis.} \ 
We evaluate major components in SSISv2 on the SOBA-testing set.
As shown in the first column in the top half of Table~\ref{table:component_analysis},
``basic'' is a network built by removing the offset vectors, class vectors, deformable MaskIoU, shadow-aware copy-and-paste augmentation, and boundary loss from SSISv2 and adopting only the segmentation loss in training.
``+ offset'' learns the offset vectors based on the ``basic'' network.
``+ class'' further considers the class vectors, the same as the model in SSIS~\cite{wang2021single}.
\wty{Table~\ref{table:component_analysis} (5-th to 11-th) rows show the new components in this work:}
``+ deformable MaskIoU'' adopts the deformable MaskIoU head to refine the predicted masks;
``+ data augm'' adopts shadow-aware copy-and-paste augmentation; and
``+ boundary loss'' leverages the boundary loss to improve the boundary accuracy. 
%
%
\wty{Table~\ref{table:component_analysis} reports the analysis results, showing that all components consistently improve the performance for most metrics and
best performance is attained when equipping all proposed components.}
%

\begin{table}[tp]
	\centering
	\caption{Evaluation on the bidirectional learning strategy.}
	\vspace*{-1mm}
	\label{table:bidirectional}
	\setlength{\tabcolsep}{7.8mm}{%
		\begin{tabular}{@{}l|cc}
			\toprule
			Strategy & $SOAP_{segm}$ & $SOAP_{bbox}$  \\ 
			\midrule
			object $\rightarrow$ shadow & 23.8  & 23.5   \\
			shadow $\rightarrow$ object &  25.6 & 23.1 \\
			\midrule
			main + associated & 26.8 & 25.8   \\
			offset pairing & 26.7 & 23.9   \\
			\midrule
			
			\textbf{SSIS} & \textbf{30.2}   & \textbf{27.2}  \\
			\bottomrule
	\end{tabular}}
	\vspace*{-0.5mm}
\end{table}

\begin{table}[tp]
	\centering
	\caption{Evaluation on the MaskIoU head strategy.}
	\vspace*{-1mm}
	\label{table:maskiou}
	\setlength{\tabcolsep}{7.8mm}{%
		\begin{tabular}{@{}l@{ }@{ }@{ }@{ }|cc}
		\toprule
		Strategy &  $SOAP_{segm}$ & $SOAP_{bbox}$ \\ \midrule
		 w/o MaskIoU head & 32.1 &  27.7\\
		 w/o deformable conv &  33.0 & 25.5  \\
		 \midrule
		\textbf{SSISv2}  &  \textbf{35.3}   & \textbf{29.0}  \\ \bottomrule
		\end{tabular}%
		}
		\vspace*{-2.5mm}
\end{table}

\vspace*{2mm}  
\noindent
{\bf Bidirectional learning strategy analysis.} \ 
\label{sec:analysis_bls}
\wty{Next, we evaluate the effectiveness of the bidirectional learning strategy.
%
First, we learn the shadow-object pairs only in one direction.}
As shown in Table~\ref{table:bidirectional}, for ``object $\rightarrow$ shadow,'' we used the architecture in Figure~\ref{fig:BRL} (a) to predict the masks of the object instances from the mask heads in the main branch, and to predict the masks of the shadow instances from the associated heads.
``shadow $\rightarrow$ object'' leverages the architecture in Figure~\ref{fig:BRL} (b) for mask prediction.
\wty{Then, we evaluate other strategies for finding the shadow-object associations.}
``main + associated'' means we use the masks predicted from the main branch and the corresponding associated branch without using the strategy in Section~\ref{sec:inference}-Inference.
\wty{``offset pairing'' means we replace the strategy in Section~\ref{sec:inference}-Inference with the learned location offset between the shadow and object instances when pairing the association.} 
%
Results show that learning the shadow-object relations from two directions with our inference strategy achieves the best performance.

\begin{table}[!t]
	\centering
	\caption{Evaluation on the boundary loss strategy.}
	\vspace*{-1mm}
	\label{table:bdloss}
	\setlength{\tabcolsep}{6.5mm}{%
		\begin{tabular}{@{}l|cc@{}}
		\toprule
		Strategy &  $SOAP_{segm}$ & \begin{tabular}[c]{@{}c@{}}Association  $AP_{segm}$ \end{tabular} \\ \midrule
		 w/o boundary loss & 33.5 & 56.8 \\
		 thick boundary loss &  34.6 & 57.3  \\
		 thin boundary loss &   34.3 & 56.8 \\
		 \midrule
		\textbf{SSISv2}  &  \textbf{35.3}   & \textbf{59.2}  \\ \bottomrule
		\end{tabular}%
		}
		\vspace*{-1mm}
\end{table}

\begin{table}[!t]
	\centering
	\caption{Evaluation on shadow-aware copy-and-paste augm.}
	\vspace*{-1mm}
	\label{table:data_aug}
	\setlength{\tabcolsep}{7.5mm}{%
		\vspace*{-1mm}
		\begin{tabular}{@{}l@{ }@{ }@{ }|cc}
			\toprule
			Strategy & $SOAP_{segm}$ & $SOAP_{bbox}$  \\ 
			\midrule
			w/o data augm.  & 33.3 & 27.5 \\
			object-only & 32.8 & 27.5 \\
			above layering &  33.8 &  27.0  \\
			multiple associations & 34.2	 & 28.3 \\
			\midrule
			\textbf{SSISv2} & \textbf{35.3}   & \textbf{29.0} \\
			\bottomrule
	\end{tabular}}
    \vspace*{-4mm}
\end{table}

\vspace*{2mm}
\noindent
\wty{\bf MaskIoU head strategy analysis.} \ 
\wty{To evaluate the effectiveness of our MaskIoU head design, we build two basic models: one by removing the MaskIoU head and the other by replacing the deformable convolution layers with na\"{I}ve convolution layers. 
Results in Table~\ref{table:maskiou} show that our MaskIoU head design with deformable convolution achieves the best performance.}

\begin{figure}[tp]
 	\centering
 	\begin{minipage}[t]{ \linewidth}
 		\includegraphics[width=0.98\linewidth]{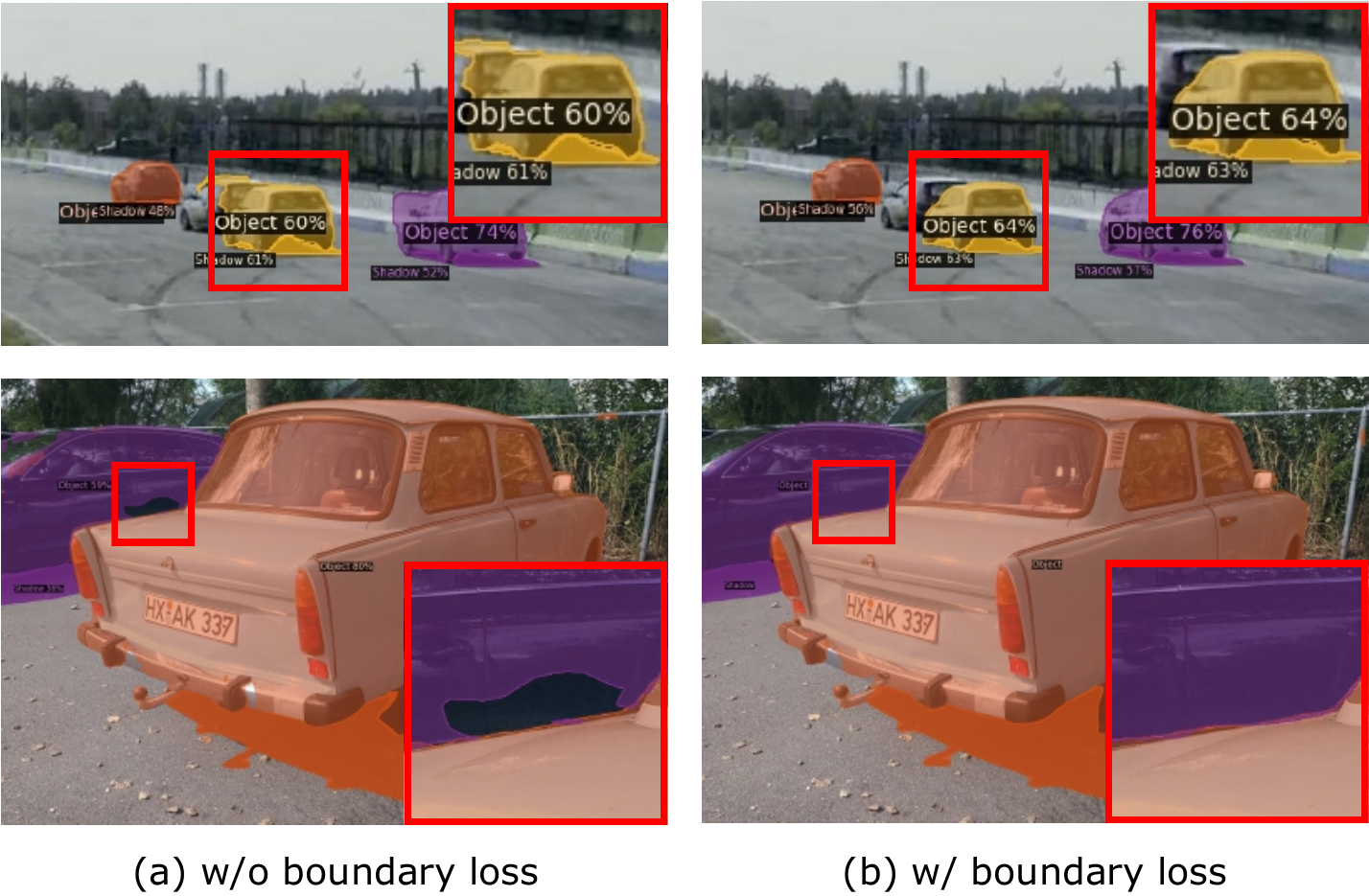}
 	\end{minipage}
 	\caption{\wty{Boundary loss analysis. (a) and (b) show the masks predicted from SSISv2 without and with boundary loss, respectively. We zoom into the regions in red boxes for better visualization. }}
 	\label{fig:boundary}
 	
 \end{figure}

\vspace*{2mm}
\noindent
\wty{\bf Boundary loss strategy analysis.} \ 
\wty{We quantitatively evaluate the effectiveness of the proposed boundary loss. 
 Table~\ref{table:bdloss} shows that both thin and thick boundary losses contribute to the performance and best performance is achieved by using both losses simultaneously; see also Figure~\ref{fig:boundary} for visual comparison results.} 

\vspace*{2mm}
\noindent
\wty{\bf Data augmentation strategy analysis.} \ 
\wty{To evaluate the effectiveness of shadow-aware copy-and-paste data augmentation, we conduct experiments with different settings (see Table~\ref{table:data_aug}), where 
	(i) ``object-only'' means we only copy and paste the object near its original position;  
	(ii) ``above layering'' means we always put the pasted shadow-object association above the original object instance;
	(iii) ``multiple associations'' means we randomly select multiple objects from the image then copy and paste the corresponding shadow-object associations to their nearby positions. 
	Table.~\ref{table:data_aug} shows that 
	(i) ``object-only'' decreases the performance compared with the baseline, since it lacks the information of the shadow instances and breaks the relations between shadows and objects;
	(ii) ``above layering'' hardly pastes the shadow naturally in front of the original object, thereby limiting the overall performance; and
	(iii) ``multiple associations'' introduces occlusions between associations, yet not as effective as SSISv2.}
 
 %
 %
 %
 %

\vspace*{2mm}  
\noindent
{\bf Discussion.}  \
SSISv2 has a strong ability of finding shadows and objects.
Yet, it is infeasible to handle some extreme scenarios, in which we cannot find another set of masks, e.g., very small shadows. In our implementation, we ignore instances that contain only one set of masks. In practice, this situation is very rare.


\begin{figure*}[tp]
	\centering
	\includegraphics[width=0.99\textwidth]{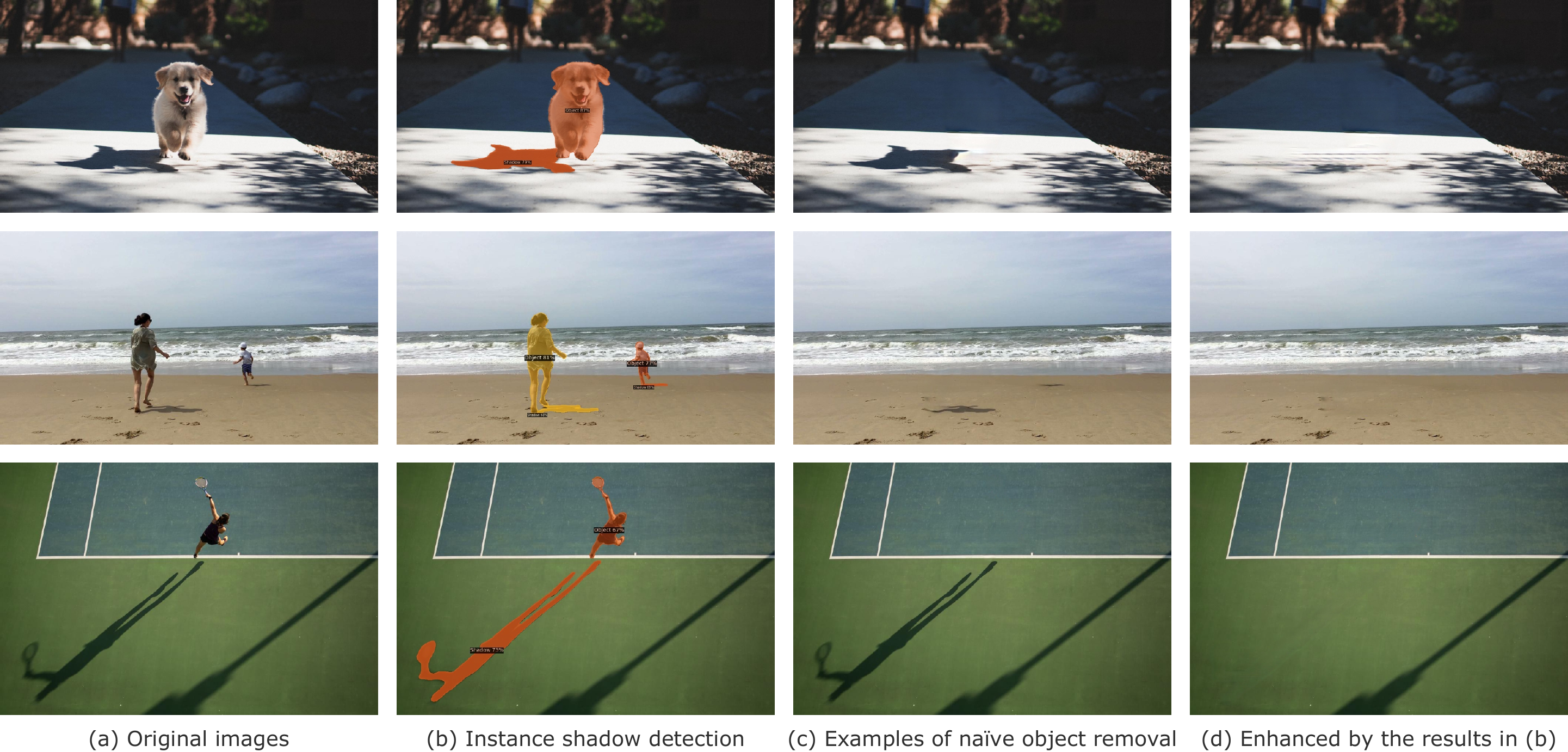}
	\vspace*{-1mm}
	\caption{Instance shadow detection enables us to easily remove objects (\eg, dog and person) with their associated shadows altogether.}
	\label{img:application2}
    \vspace*{-1.5mm}
\end{figure*}

\begin{figure}[tp]
	\centering
	\includegraphics[width=0.95\linewidth]{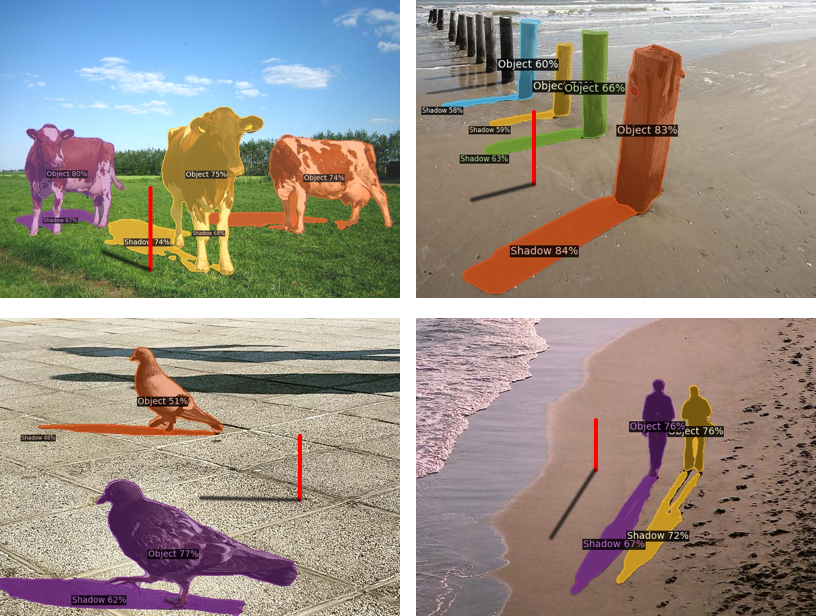} 
	\caption{\wty{We estimate the light direction and incorporate a virtual red post into each image with a simulated shadow, following~\cite{Lalonde2009illum}.}}
	\label{img:application1}
	\vspace*{-3mm}
\end{figure}

\section{Applications}

%

\wty{Below, we present application scenarios to demonstrate the applicability of the results produced by our SSISv2.}

\vspace*{2mm}
\noindent
{\bf Light direction estimation.} \
%
%
\wty{Instance shadow detection promotes 2D light direction estimation in the image planes.}
For instance, we can connect the bounding box centers of the shadow and object instances in each shadow-object association pair as the estimated light direction.
%
%
%
Figure~\ref{img:application1} shows some example results, for which we adopt the estimated light directions to render virtual red posts with simulated shadows on the ground.
From the results, we can see that the virtual shadows with the red posts~\cite{Lalonde2009illum} look consistent with the real shadows cast by the other objects, thus showing the applicability of our detection results of our method.
%

%
 


\begin{figure}[tp]
	\centering
	\includegraphics[width=0.98\linewidth]{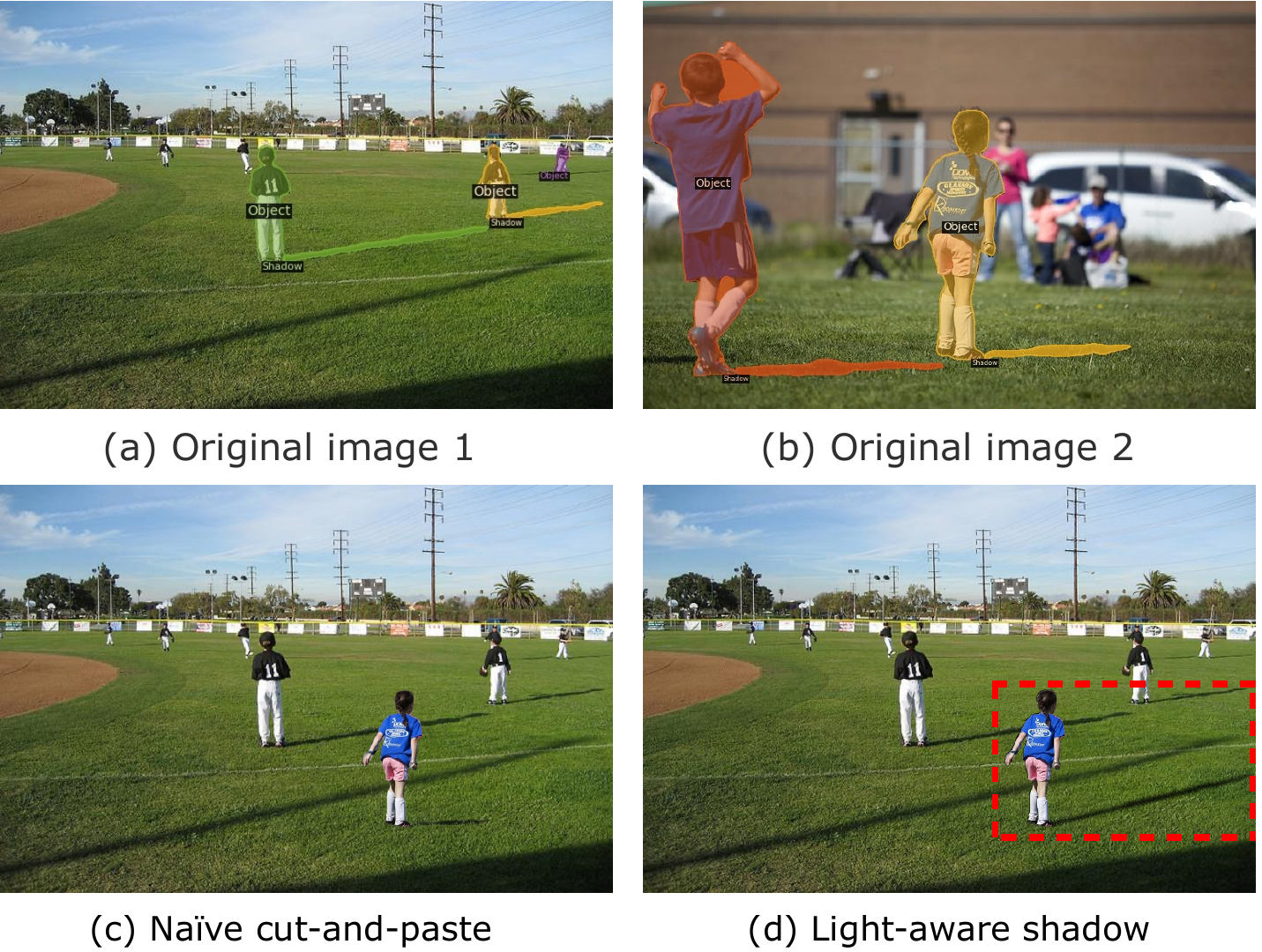}
	\caption{When we cut-and-paste objects from one photo to the other, instance shadow detection results enable us not only to extract object and shadow instances together but also to adjust the shadow shape according to the estimated light direction.}
	\label{img:application3}
	\vspace*{-3mm}
\end{figure}



\vspace*{2mm}
\noindent
{\bf Photo editing.} \
%
Another application is photo editing, in which we can remove object instances together with their associated shadows.
%
Yi~\etal~\cite{Yi_2020_CVPR} developed an image in-painting method for automatically removing specific objects by a given corresponding mask. 
With the results of instance segmentation methods, we can remove specific objects but leave shadows cast by the objects on the ground; see Figure~\ref{img:application2} (c).
%
With the help of our instance shadow detection results (Figure~\ref{img:application2} (b)), we can remove the objects with their shadows altogether, as shown in Figure~\ref{img:application2} (d).

Further, we can efficiently transfer an object together with its shadow from one photo to another.
%
Figure~\ref{img:application3} shows an example, in which we remove the girl together with her shadow from (b) and paste them together onto (a) in a smaller size.
Clearly, if we simply paste the girl and shadow to (a), the shadow is not consistent with the real shadows in the target photo; see (c).
\wty{Thanks to instance shadow detection, which outputs individual masks for objects together with their associated shadow instances, as well as the estimated 2D light direction.}
So, we can achieve light-aware photo editing by using the estimated light directions in both photos to adjust the shadow images when transferring the girl object from one photo to the other; see (d).




\section{Conclusion}
\label{sec:conclusion}

This paper presents instance shadow detection, targeting to predict shadow instances, object instances, and their relations. 
To approach this task, we first prepare a new dataset and a new evaluation metric.
Our dataset contains 1,100 images with labeled masks of 4,262 pairs of shadow instances, object instances, and shadow-object associations, while the evaluation metric promotes quantitative evaluation of instance shadow detection performance.
We also design a new single-stage fully-convolutional network for instance shadow detection by directly learning the relation between shadow instances and object instances in an end-to-end manner.
Further, we propose the bidirectional relation learning module, the deformable maskIoU head, and the shadow-aware copy-and-paste augmentation to improve the detection performance.
Finally, we show the superiority of our method on the benchmark dataset and demonstrate the applicability of our method on light direction estimation and photo editing.

In the future, we plan to improve the performance of instance shadow detection by exploring the knowledge from the existing data prepared for other relevant vision tasks,~\eg, shadow detection and instance segmentation, from synthetic data generated by computer graphic techniques and from unlabeled data downloaded from the Internet.
Also, we plan to explore more applications based on the shadow-object association results.

\ifCLASSOPTIONcompsoc
  \section*{Acknowledgments}
\else
  \section*{Acknowledgment}
\fi

This work was supported by the project MMT-p2-21 of the Shun Hing Institute of Advanced Engineering, The Chinese University of Hong Kong, and the Hong Kong Research Grants Council under General Research Fund (CUHK 14201620 \& CUHK 14201321).


\ifCLASSOPTIONcaptionsoff
  \newpage
\fi



%

{\small
	\bibliographystyle{IEEEtran}
	\bibliography{reference}
}



%




\end{document}